\documentclass[journal,onecolumn,10pt,draftclsnofoot]{IEEEtran}
\usepackage[T1]{fontenc}
\usepackage{cite}
\usepackage{amsmath,amssymb}
\usepackage{algorithm}
\usepackage{graphicx}
\usepackage{textcomp}
\usepackage{xcolor}
\usepackage{algorithm} 
\usepackage{algpseudocode} 
\usepackage{dsfont}
\usepackage{changepage}
\usepackage{setspace}
\usepackage{pdfpages}
\usepackage{mathrsfs}
\usepackage{diagbox}
\usepackage{multirow}
\usepackage{lscape}
\usepackage{rotating}
\usepackage{subfig}
\usepackage{lipsum}
\usepackage[normalem]{ulem}
\usepackage{booktabs}

\newtheorem{thm}{Theorem}

\newtheorem{lem}[thm]{Lemma}

\newtheorem{remark}{Remark}

\newcommand\E{\mathbb{E}}

\newcommand\Cov{\operatorname{Cov}}
\newcommand\Tr{\operatorname{Tr}}

\begin{document}

\title{Traversing Distortion-Perception Tradeoff using a Single Score-Based Generative Model}

\author{
  Yuhan Wang, Suzhi Bi,
  Ying-Jun Angela Zhang, and Xiaojun Yuan
  \thanks{The work of Yuhan Wang and Ying-Jun Angela Zhang is supported in part by the General Research Fund (project number 14202421, 14214122, 14202723), Area of Excellence Scheme grant (project number AoE/E-601/22-R), and NSFC/RGC Collaborative Research Scheme (project number CRS\_HKUST603/22), all from the Research Grants Council of Hong Kong. The work of Suzhi Bi is supported in part by the Natural Science Foundation of China under Project 62271325; in part by Guangdong Basic and Applied Basic Research Foundation under Project 2024B1515020089; in part by Shenzhen Science and Technology Program under Project 20220810142637001. The work of Xiaojun Yuan is supported in part by Sichuan Science and Technology Program under Grant 2024ZYD0036 and in part by National Key Laboratory of Wireless Communications Foundation under Grant IFN20230204. \emph{(Corresponding author: Suzhi Bi.)}}
  \thanks{Yuhan Wang and Ying-Jun Angela Zhang are with the Department of Information Engineering, The Chinese University of Hong Kong, Hong Kong (e-mail: wy023@ie.cuhk.edu.hk; yjzhang@ie.cuhk.edu.hk).}
  \thanks{Suzhi Bi is  with the College of Electronic and Information Engineering, Shenzhen University, Shenzhen 518060,
  China (e-mail: bsz@szu.edu.cn).}
  \thanks{Xiaojun Yuan is with the National Key Laboratory of Wireless Communications, University of Electronic Science and Technology of China, Chengdu
  611731, China (e-mail: xjyuan@uestc.edu.cn).}}



\maketitle

  \begin{abstract}
    The distortion-perception (DP) tradeoff reveals a fundamental conflict between distortion metrics (e.g., MSE and PSNR) and perceptual quality. Recent research has increasingly concentrated on evaluating denoising algorithms within the DP framework. However, existing algorithms either prioritize perceptual quality by sacrificing acceptable distortion, or focus on minimizing MSE for faithful restoration. When the goal shifts or noisy measurements vary, adapting to different points on the DP plane needs retraining or even re-designing the model. Inspired by recent advances in solving inverse problems using score-based generative models, we explore the potential of flexibly and optimally traversing DP tradeoffs using a single pre-trained score-based model. Specifically, we introduce a variance-scaled reverse diffusion process and theoretically characterize the marginal distribution. We then prove that the proposed sample process is an optimal solution to the DP tradeoff for conditional Gaussian distribution. Experimental results on two-dimensional and image datasets illustrate that a single score network can effectively and flexibly traverse the DP tradeoff for general denoising problems.
\end{abstract}


\section{Introduction}\label{sec:intro}

\IEEEPARstart{I}{n recent} years, we have witnessed rapid progress in image restoration algorithms, especially deep learning-based implementations that have shown remarkable achievements in solving general inverse problems. How to evaluate the performance of emerging algorithms is a crucial yet complicated problem. Traditional full reference \emph{distortion} metrics, such as Mean Square Error (MSE), focus on the pixel-level accuracy between the original image $X$ and its reconstruction $\hat{X}$. On the other hand, \emph{perceptual quality}, referring to the degree to which an image looks natural rather than algorithmically generated \cite{moorthy2011blind_perceptual}, is also an important measure. It has been demonstrated that perceptual quality could be associated with the distance between the distributions of natural images $p_X$ and generated images $p_{\hat{X}}$ \cite{PD-tradeoff,RethinkingRDP,CDP}. 

Improvements in distortion measures do not necessarily lead to enhancements in perceptual quality. In fact, \cite{PD-tradeoff} has demonstrated that there exists a tradeoff between distortion and perception, termed as distortion-perception (DP) tradeoff. Mathematically, it can be modeled as 
\begin{align*}
    D(P) = 
    &\min_{p_{\hat{X}|Y}}\mathbb{E}[\Delta(X,\hat{X})],\\
    \quad\text{s.t.}~ &d(p_X,p_{\hat{X}}) \leq P,
\end{align*}
where $Y$ is the degraded observation, $\Delta:\mathcal{X}\times\hat{\mathcal{X}}\to\mathbb{R}^+$ is a full reference distortion function (e.g., such as square-error), and $d(\cdot, \cdot)$ is some divergence between probability distribution, such as Wasserstein distance \cite{Invitation_Wasserstein_space} or Kulback-Leibler (KL) divergence. 

Since the introduction of the DP tradeoff, more and more algorithm designs have focused on performance evaluation on the DP plane for specific denoising tasks, e.g., image deblurring \cite{PSCGAN,Deblur-Stoch-Refine} and super-resolution \cite{Lim_2017_CVPR_Workshops,wang2018esrgan,marinescu2021bayesian-DNN}. However, most algorithms tend to seek better perceptual quality at the cost of distortion, or focus on optimizing MSE, a distortion measure, to ensure restoration fidelity. When the restoration objective shifts between distortion and perception, we need to retrain or even re-design the network, which can be computationally intensive. In many practical scenarios, it is significant to flexibly and effectively fulfill diverse task objectives or user demands with a \emph{single} model at inference time. 

There have been explorations into theories and algorithms to traverse the DP tradeoff in inference time \cite{PSCGAN,DP-tradeoff_Wasserstein_Freirich2021}. Specifically, in \cite{PSCGAN}, the authors developed a framework based on conditional generative adversarial networks (CGAN) with an additional penalty on posterior expectation. By averaging different numbers of samples or adjusting the noise level injected into the generator, this CGAN-based framework can access various points along the tradeoff. However, using multiple image samples and averaging them may not be efficient during inference, and the method is not provably optimal. Furthermore, the training process relies on pairs of clean and noisy data. Thus, retraining is required for different noise levels and measurements. In \cite{DP-tradeoff_Wasserstein_Freirich2021}, the DP tradeoff is studied in Wasserstein space. Theoretical results show that estimators on the DP curve can be constructed by linearly combining two estimators at the extremes: an optimal MSE minimizer and a perfect perception sampler with minimum distortion. This idea is further applied in burst restoration \cite{brustDP}. In practice, a method that achieves a relatively low MSE, alongside another method that achieves good perceptual quality, can serve as approximations of these two extreme estimators for a specific task. Nonetheless, different models must be deployed to address various measurements, rendering this method less flexible to general inverse problems.

Nonetheless, the optimal estimators at the two extremes may not be available in practice. Additionally, neither of the two aforementioned methods can be flexibly applied to general inverse problems, as they require distinct models for varying measurements. Inconsistencies in noise levels between training data and observations would lead to invalid reconstruction.

Recent research has demonstrated the generative capability of score-based diffusion models  \cite{DDPM,song2019gradient,kawar2022denoising,song2021scorebased, song2021DDIM, bortoli2021diffusion_Schrodinger_Bridge_Score} in tackling general inverse problems using a single score network \cite{chung2022DiffusionManifold,chung2023DPS,wang2023zeroshotDDNM,song2023pseudoPGDM,DiffPIR,xu2024provablyDPS_Plugin}.
Score-based diffusion models learn the prior distribution of the data $\mathbf{x}$ by training a score network to match the gradient of the logarithm density $\nabla_{\mathbf x}\log p(\mathbf{x})$, referred to as the \emph{score}. After observing a noisy measurement $\mathbf{y}$, sampling from posteriors involves approximating the conditional score $\nabla_{\mathbf x}\log p(\mathbf{x}|\mathbf{y})$. Theoretical insights provided in \cite{Xue2024_scorebased_variational_infer} also reveal the potential for recovering Minimum Mean Square Error (MMSE) estimation by propagating the mean of the reverse diffusion process. Inspired by these advancements, we explore the potential of traversing DP tradeoffs for different tasks using a single score network. Our main contributions are summarized below:
\begin{itemize}
    \item  First, we propose a variance-scaled reverse diffusion process and theoretically characterize the marginal distributions produced by this novel reverse process. It is demonstrated that the mean of our proposed sampling converges towards the MMSE point while the marginal covariance is scaled by a scaling factor. By tuning a parameter that dictates the variance of reverse sampling, we can flexibly navigate from the MMSE point to the posterior distribution, where the two extremes are achieved by setting the parameter to zero and one, respectively.
    \item Subsequently, we show that the reconstruction obtained from the variance-scaled reverse sampling represents the optimal solution to the conditional DP tradeoff for multivariate Gaussian distributions, assessed through MSE and Wasserstein-2 distance metrics. 
    \item Finally, we validate our methodology through experiments conducted on various twoßdimensional datasets and a real-world image dataset. With a single pre-trained score network, we can traverse the DP tradeoff across different inverse problems with varying noise levels. The results indicate that the variance-scaled reverse diffusion process achieves a more complete empirical DP tradeoff curve and better MSE than the GAN-based method and other inverse problem solvers, demonstrating the effectiveness and flexibility of the proposed framework.
\end{itemize}

\textbf{Notations:} For a random variable $X$ denoted by a capital letter, we use small letter $\mathbf x$ to denote its realizations, and use $p_X(\mathbf x$) to denote the distribution over its alphabet $\mathcal{X}$. When there is no ambiguity, the distribution can be abbreviated as $p(\mathbf x)$. For a discrete sequence of random variables $X_0, X_1, \cdots, X_T$, we abbreviate them as $\{X_k\}_{k=0}^{T}$ and use $\mathbf x_{0:T}$ to denote their realizations. 
The expectation and conditional expectation of $X$ given $Y=\mathbf y$ is $\E_{p(\mathbf{x})}[X]$ and $\E_{p(\mathbf{x}|\mathbf{y})}[X]$. We use $\Cov_{p(\mathbf{x})}[X]$ and $\Cov_{p(\mathbf{x}|\mathbf{y})}[X]$ to represent the covariance and conditional covariance of $X$ given $Y=\mathbf y$. Matrices are denoted by uppercase boldface letters (e.g., $\mathbf \Sigma$). $\Tr(\mathbf \Sigma)$ and $\mathbf \Sigma^{-1}$ represent the trace and inverse.

\section{Backgrounds}\label{sec: backgrounds}
\subsection{Score-based generative model}
Score-based generative models or diffusion models define the generative process as the reverse of a diffusion process. Specifically, the diffusion process (i.e., forward process) is a Markov chain $\{X_k\}_{k=0}^T$, with joint distribution 
\begin{align}
    p(\mathbf x_{0:T}):=p(\mathbf x_0)\prod_{k=0}^{T-1}p(\mathbf x_{k+1}|\mathbf x_{k}), \label{forward_prob}
\end{align}
which gradually diffuses the data $X_0\in\mathbb{R}^{d}$ with distribution $p(\mathbf{x}_0) \triangleq p_{\text{data}}$. Variance preserving (VP) diffusion \cite{DDPM} adopts $p(\mathbf x_{k+1}|\mathbf x_{k})\triangleq\mathcal{N}(\mathbf x_{k+1};\sqrt{1-\beta_{k+1}}\mathbf x_{k},\beta_{k+1} \mathbf I)$ with monotonically increasing variance schedule $\beta_T\geq \beta_{T-1}\geq \cdots \geq \beta_{0}=0$. When $T\to\infty$, $\mathbf{x}_T$ converges to an isotropic Gaussian distribution  $\mathcal{N}(\mathbf{0}, \mathbf{I})$. 

To generate a sample following the data distribution $p(\mathbf{x}_0)$, we can start from the standard Gaussian $p_{\theta}(\mathbf x_T):=\mathcal{N}(\mathbf{0}, \mathbf{I})$ and follow the reverse process
\begin{align}
    p_{\theta}(\mathbf{x}_{0:T}):=p_{\theta}(\mathbf x_T)\prod_{k=0}^{T-1}p_{\theta}(\mathbf x_{k}|\mathbf x_{k+1}), \label{reverse_prob}
\end{align}
with transitions $p_{\theta}(\mathbf x_{k}|\mathbf x_{k+1})\sim\mathcal{N}(\mu_{\theta}(\mathbf{x}_k, k), \mathbf{\Sigma}_{\theta}(\mathbf{x}_k, k))$ parameterized by $\theta$. It will be shown later that the reverse sampling is closely related to the \emph{score function} for each step, i.e., $\nabla_{\mathbf{x}_k}\log p(\mathbf{x}_k),$ for $k=0,1,\cdots, T$.

In continuous-time, we can describe the diffusion process $(X_t)_{t\in[0,T_c]}$, $X_t\in\mathbb{R}^d$ with a stochastic differential equation (SDE) \cite{song2021scorebased} 
\begin{align}
    dX_t=-\frac{1}{2}\beta(t)X_tdt+\sqrt{\beta(t)}dW_t, ~X_{0}\sim p_{\text{data}}, \label{forward_SDE}
\end{align}
where $(W_t)_{t\in[0,T_c]}$ is a standard Brownian motion, and $\beta(t)$ is the noise schedule. Let $(p_t)_{t\in[0:T_c]}$ be the path distribution associated with \eqref{forward_SDE}. To perform data sampling following the distribution $p_{\text{data}}$, we can reverse the SDE \eqref{forward_SDE} and apply discretization.
According to \cite{ANDERSON1982_reverseSDE} and \cite{song2021scorebased}, the reverse SDE corresponding to \eqref{forward_SDE} is 
\begin{align}
    d\overleftarrow{X}_{t}=\Big[\frac{1}{2}&\beta(T_c\!-\!t)\overleftarrow{X}_{t}+\beta(T_c\!-\!t)\nabla\log p_{T_c\!-t}(\overleftarrow{X}_{t})\Big]dt +\sqrt{\beta(T_c\!-\!t)}d\tilde{W}_t, ~\overleftarrow{X}_0\sim p_{T_c},\label{reverse_SDE}
\end{align} 
$\!\!\text{where}$ $(\tilde W_t)_{t\in[0,T_c]}$ is another standard Brownian motion. The reverse SDE produces a time-reverse process $(\overleftarrow{X}_t)_{t\in[0,T_c]}$ where $\overleftarrow{X}_t$ has the same distribution with $X_{T_c-t}$. 
Note that the drift function in \eqref{reverse_SDE} depends on the score function $\nabla_{\mathbf{x}_t}\log p_{t}(\mathbf{x}_t)$, which can be approximated by a time-aware neural network trained with denoising score matching \cite{Vincent_score_matching}.

\subsection{Score-based model for posterior sampling}
In many application scenarios, we may observe a noisy version of the measurement of the data $X_0$, given by
\begin{align*}
    Y = \mathcal{A}(X_0) + N,
\end{align*}
where $\mathcal{A}(\cdot):\mathbb{R}^d\to\mathbb{R}^n$ is a measurement operator and $N\in\mathbb{R}^n$ is the measurement noise with $N\sim\mathcal{N}(\mathbf{0},\sigma_n^2\mathbf{I})$. 

Score-based generative models have shown powerful capability in solving general inverse problems \cite{kawar2022denoising,chung2023DPS,song2023pseudoPGDM,DiffPIR,Xue2024_scorebased_variational_infer,xu2024provablyDPS_Plugin}. Leveraging the diffusion model as the prior, sampling from the reverse posterior distribution
\begin{align}
    p(\mathbf{x}_{0:T}|\mathbf{y}):=p(\mathbf x_T|\mathbf{y})\prod_{k=0}^{T-1}p(\mathbf x_{k}|\mathbf x_{k+1},\mathbf{y}) \label{reverse_post_prob}
\end{align}
requires the knowledge of conditional scores $\nabla_{\mathbf{x}_k}\log p(\mathbf{x}_k|\mathbf y)$, which can be expressed as 
    $\nabla_{\mathbf{x}_k}\log p(\mathbf{x}_k|\mathbf y) = \nabla_{\mathbf{x}_k}\log p(\mathbf{x}_k) + \nabla_{\mathbf{x}_k}\log p(\mathbf y|\mathbf{x}_k)$.
The first term can be approximated by a score network $s_{\theta}(\mathbf{x}_k,k)$ \cite{DDPM,song2021scorebased}, and different methods have been proposed to estimate the second term $\nabla_{\mathbf{x}_k}\log p(\mathbf y|\mathbf{x}_k)$ \cite{chung2023DPS,song2023pseudoPGDM}. In each step the conditional sampling process in VP diffusion is \cite{chung2023DPS,song2023pseudoPGDM}
\begin{align}
    \mathbf x_{k-1}&\leftarrow\frac{1}{\sqrt{\alpha}_k}(\mathbf x_k+(1\!-\!\alpha_k)\nabla_{\mathbf x_k}\log p(\mathbf x_k|\mathbf y)) +\tilde\sigma_k\mathbf z,\label{conditional_sampling_step}
    \\
    &\approx \frac{1}{\sqrt{\alpha}_k}(\mathbf x_k+(1\!-\!\alpha_k)(s_{\theta}(\mathbf{x}_k,k)\! +\! \nabla_{\mathbf{x}_k}\log p(\mathbf y|\mathbf{x}_k)))+\tilde\sigma_k\mathbf z,\notag
\end{align} 
$\!\!\text{where}$ $\mathbf z\sim\mathcal{N}(0,\mathbf I)$ and $\tilde\sigma_k^2$ is the approximated variance of the reverse posterior distribution.

Meanwhile, \cite{Xue2024_scorebased_variational_infer} proposed to directly approximate the reverse conditional $p(\mathbf x_{k}|\mathbf x_{k+1},\mathbf{y})$ and propagate the mean in each step. Specifically, when $T\to\infty$, the mean of $p(\mathbf x_{k}|\mathbf x_{k+1},\mathbf{y})$ in VP-diffusion is given by \cite{Xue2024_scorebased_variational_infer} 
\begin{align}
    &\boldsymbol{\mu}_k(\mathbf x_{k+1}, \mathbf y)=\mathbf U_{k}\mathbf x_{k+1}+\mathbf V_{k}\mathbb E_{p(\mathbf x_0|\mathbf y)}[X_0], \label{reverse_mean}
\end{align}
where
\begin{align*}
    &\mathbf U_{k} := \sqrt{\alpha_{k+1}}\big((1-\bar{\alpha}_{k})\mathbf I+\bar{\alpha}_{k}\text{Cov}_{p(\mathbf x_0|\mathbf y)}[X_0]\big)\cdot\big((1-\bar{\alpha}_{k+1})\mathbf I+\bar{\alpha}_{k+1}\text{Cov}_{p(\mathbf x_0|\mathbf y)}[X_0]\big)^{-1},\\
    &\mathbf V_{k} := \sqrt{\bar{\alpha}_k}\big((1\!-\!\alpha_{k\!+\!1})\mathbf I\big)\big((1\!-\!\bar{\alpha}_{k\!+\!1})\mathbf I\!+\!\bar{\alpha}_{k\!+\!1}\text{Cov}_{p(\mathbf x_0|\mathbf y)}[X_0]\big)^{\!\!-1}\!\!\!.
\end{align*}
Here $\alpha_k=1-\beta_k$ and $\bar{\alpha}_k=\Pi_{i=0}^k\alpha_i$ in VP-diffusion.
The authors proved that by propogating the mean at each reverse step (i.e., $\boldsymbol{\mu}_T\to \boldsymbol{\mu}_{T-1}(\mathbf x_{T}=\boldsymbol{\mu}_T, \mathbf y)\to\cdots\to \boldsymbol{\mu}_{0}(\mathbf x_{1}=\boldsymbol{\mu}_1, \mathbf y)$), the end point $\boldsymbol{\mu}_{0}(\mathbf x_{1}=\boldsymbol{\mu}_1, \mathbf y)$ represents the MMSE estimator $\mathbb E_{p(\mathbf x_0|\mathbf y)}[X_0]$ when $\bar\alpha_T\to 0$. The proof of reverse mean propagation converging to the MMSE point \cite{Xue2024_scorebased_variational_infer} and its connection to conditional score are included in Appendix \ref{Appendix_approx_reverse_condi}.

\subsection{Connection between reverse mean and conditional score}
The mean derived in \eqref{reverse_mean} can be viewed as an approximation of the mean in \eqref{conditional_sampling_step}. Specifically, when $T\to\infty$, $p(\mathbf{x}_k|\mathbf y)$ can be approximately expressed in the form of a Gaussian distribution. We can show that (see details in Appendix \ref{Appendix_approx_reverse_condi})
\begin{align*}
    \frac{1}{\sqrt{\alpha}_k}&\!(\mathbf x_k+(1\!-\!\alpha_k)\nabla_{\mathbf x_k}\!\log p(\mathbf x_k|\mathbf y))=\mathbf U_{k}\mathbf x_{k+1}+\mathbf V_{k}\mathbb E_{p(\mathbf x_0|\mathbf y)}[X_0].
\end{align*}

Theoretically, when $\tilde{\sigma}_k=0$ in each step \eqref{conditional_sampling_step}, the endpoint of the reverse process converges to the MMSE estimator, which achieves the best possible performance on distortion. When $\tilde{\sigma}_k$ is the true posterior variance, the reverse process samples from the posterior distribution $p(\mathbf{x}_{0:T}|\mathbf{y})$, which results in a reconstruction with perfect perception measured in conditional distribution. The above observations motivate us to \textbf{bridge the two extreme points with the score-based generative model}  by tuning the scale of $\sigma_k$ in the reverse process. It remains to be answered whether the bridging can provide us with an optimal and flexible estimator to traverse along the distortion-perception plane.

\section{Traversing DP Tradeoff with Scaled Reverse Diffusion}
In this section,  we first propose a novel reverse sampling method based on a variance-scaled version of the joint inference distribution \eqref{reverse_post_prob} and theoretically derive the mean and variance of the \emph{marginal} distribution. Then, we prove that the resulting end distribution of the variance-scaled reverse sampling provides an optimal DP tradeoff with respect to MSE and Wasserstein-2 distance. 

\subsection{Reverse mean and variance}
Consider the joint inference distribution with true mean but scaled variance, i.e., 
\begin{align}
    p_{\lambda}(\mathbf x_{k}|\mathbf x_{k+1},\mathbf y):=\mathcal{N}\Big(\boldsymbol{\mu}_k(\mathbf x_{k+1}, \mathbf y), ~\lambda \mathbf C_k\Big), \label{reverse_dist_lambda}
\end{align}
where the expectation $\boldsymbol\mu_k(\mathbf x_{k+1}, \mathbf y)$ is given in \eqref{reverse_mean}, and the true covariance $\mathbf C_k$ is given by \cite{Xue2024_scorebased_variational_infer}
\begin{align}
    &\mathbf C_k := \frac{\beta_{k+1}}{1-\beta_{k+1}}\big((1-\bar{\alpha}_k)\mathbf I+\bar{\alpha}_k\text{Cov}_{p(\mathbf x_0|\mathbf y)}[X_0]\big)\cdot\Big(\big(\frac{\beta_{k+1}}{1-\beta_{k+1}}+1-\bar{\alpha}_k\big)\mathbf I+\bar{\alpha}_k\text{Cov}_{p(\mathbf x_0|\mathbf y)}[X_0]\Big)^{-1}\!\!\!.\label{reverse_cov}
\end{align}
We have the following theorem to characterize the marginal distribution given by the variance-scaled version of the joint inference distribution. 

\begin{thm}\label{thm_reverse_mean_var}
    For VP-diffusion and $0\leq \lambda\leq 1$, consider the joint inference distribution given by
    \begin{align}
        p_{\lambda}(\mathbf x_{0:T}|\mathbf y)=p_{\lambda}(\mathbf x_T|y)\prod_{k=0}^{T-1}p_{\lambda}(\mathbf x_{k}|\mathbf x_{k+1},\mathbf y) \label{Eq_joint_inference_dist}
    \end{align}
    where $p_{\lambda}(\mathbf  x_T|\mathbf y)=\mathcal{N}(0,\mathbf I)$ and $p_{\lambda}(\mathbf x_{k}|\mathbf x_{k+1},\mathbf y)$ is given by \eqref{reverse_dist_lambda}.
    Then, the corresponding margin has the distribution $p_{\lambda}(\mathbf x_k|\mathbf y) = \mathcal{N}(\boldsymbol{\mu}_k^{\lambda}, \mathbf \Sigma_k^{\lambda})$, where
    \begin{align*}
        \boldsymbol{\mu}_{k}^{\lambda} &= \Big(\!\mathbf U_{k}\big(\mathbf U_{k+1}(\cdots (\mathbf U_{T-2}\mathbf V_{T-1}+\mathbf V_{T-2})\cdots)+\mathbf V_{k+1}\big)+\mathbf V_{k}\Big)\\
        &=\sqrt{\bar{\alpha}_k}(1-\alpha_{k\!+\!1}\cdots\alpha_T)\big((1\!-\!\bar{\alpha}_T)\mathbf I\!+\!\bar{\alpha}_T\text{Cov}_{p(\mathbf x_0|\mathbf y)}[X_0]\big)^{\!-1}\E_{p(\mathbf x_0\!|\mathbf y\!)}[X_0],\\ 
        \mathbf \Sigma_{k}^{\lambda}
        &=\mathbf \Sigma_{k}\Big(\lambda \mathbf I+(1-\lambda)\alpha_{k+1}\alpha_{k+2}\cdots\alpha_T\mathbf \Sigma_{T-1}^{-1}\mathbf \Sigma_{k}\Big),
    \end{align*}
    where $\mathbf \Sigma_k$ is the covariance of $p(\mathbf x_k|\mathbf y)$ in VP diffusion. In particular, when $\bar\alpha_T\to 0$, the variance of $p_{\lambda}(\mathbf x_0|\mathbf y)$ is 
    \begin{align*}
        \mathbf \Sigma_{0}^{\lambda}&=\mathbf \Sigma_{0}\Big(\lambda \mathbf I+(1-\lambda)\bar{\alpha}_T\mathbf \Sigma_{T-1}^{-1}\mathbf \Sigma_{0}\Big)
        \to \lambda\text{Cov}_{p(\mathbf x_0|\mathbf y)}[X_0],
    \end{align*}
    and the mean is
    \begin{align*}
        \mu_0^{\lambda}=\sqrt{\bar{\alpha}_0}\big((1-\bar{\alpha}_T)\mathbf I\big)&\big((1-\bar{\alpha}_T)\mathbf I+\bar{\alpha}_T\text{Cov}_{p(\mathbf x_0|\mathbf y)}[X_0]\big)^{-1}\cdot\mathbb E_{p(\mathbf x_0|\mathbf y)}[X_0]\to \mathbb E_{p(\mathbf x_0|\mathbf y)}[X_0].
    \end{align*}
\end{thm} 

\begin{IEEEproof}
    See the details in Appendix \ref{Appendix_Proof_of_scaled_mean_var}.
\end{IEEEproof}

The above theorem shows that the proposed variance-scaled reverse diffusion process will lead to a scaled marginal distribution at each step, which finally results in the posterior mean and scaled posterior variance of the original data distribution at the end of the reverse process. This seemingly intuitive result actually provides the optimal solution to the conditional distortion-perception tradeoff for  multivariate Gaussian distribution.

\subsection{DP tradeoff for conditional Gaussian case}
Consider the $d$-dimensional source $X\in\mathbb{R}^d$ with distribution $p(\mathbf{x})$. Let $Y$ be the noisy version of the source data. Denote the conditional expectation and variance given an observation $\mathbf y$ as $\boldsymbol \mu_{\mathbf y} = \E_{p(\mathbf x|\mathbf y)}[X]$ and $ \mathbf \Sigma_{\mathbf y} = \Cov_{p(\mathbf x|\mathbf y)}[X]$, respectively. The goal of signal restoration is to find a good reconstruction $\hat{X}$ based on the observed $Y$.

From the optimization perspective, given a specific observation $Y=\check{\mathbf{y}}$, we can define the \emph{conditional} distortion-perception function as
\begin{align*}
    D(P) = \min_{p_{\hat{X}|Y}(\hat{\mathbf{x}}|\check{\mathbf y})}&\mathbb{E}_{p_{X,\hat{X}|Y}(\mathbf{x},\hat{\mathbf{x}}|\check{\mathbf y})}[||X-\hat{X}||_2^2]\\
    \text{s.t.}~ &d(p_{X|Y}(\mathbf{x}|\check{\mathbf y}), p_{\hat{X}|Y}(\hat{\mathbf{x}}|\check{\mathbf y}))\leq P,
\end{align*}
where $X,Y$ and $\hat{X}$ form a Markov chain.

\begin{thm} \label{thm_optimality_condi_Gaussian}
    Consider the source $X$ and an observation $\check{\mathbf y}$ that satisfy $p_{X|Y}(\mathbf x|\check{\mathbf y})\sim\mathcal{N}(\boldsymbol\mu_{\check{\mathbf y}}, \mathbf \Sigma_{\check{\mathbf y}})$. The optimal distortion-perception tradeoff given $\check{\mathbf y}$ with MSE and Wasserstein-2 distance is  
    \begin{align}
        D(P)\!=\!\begin{cases}
            \Tr(\mathbf \Sigma_{\check{\mathbf y}})\!+\!\big(\sqrt{\Tr(\mathbf \Sigma_{\check{\mathbf y}})}\!-\!P\big)^2, ~\text{for } P\!\leq \!\sqrt{\Tr(\mathbf \Sigma_{\check{\mathbf y}})}\\
            \Tr(\mathbf \Sigma_{\check{\mathbf y}}),~\text{for }  P>\sqrt{\Tr(\mathbf \Sigma_{\check{\mathbf y}})}. 
        \end{cases}\label{DP_optimal}
    \end{align}
    Meanwhile, the optimal tradeoff can be achieved by sampling from the joint inference distribution \eqref{Eq_joint_inference_dist} in Theorem \ref{thm_reverse_mean_var} and traversing different $0\leq \lambda\leq 1$.
\end{thm}
\begin{IEEEproof}
    See details in Appendix \ref{Appendix_Proof_Optimality}.
\end{IEEEproof}

 Theorem \ref{thm_optimality_condi_Gaussian} explicitly characterizes the optimal DP tradeoff for conditional Gaussian distribution. When $P=0$ (i.e., perfect perceptual quality is expected), the best achievable MSE is $2\Tr(\mathbf \Sigma_{\check{\mathbf y}})$. As $P$ increases, the optimal MSE gradually converges to the MMSE value $\Tr(\mathbf \Sigma_{\check{\mathbf y}})$. The proof in Appendix \ref{Appendix_Proof_Optimality} also shows that the variance-scaled reverse diffusion process can attain the entire optimal DP curve for conditional Gaussian distribution. Although the actual data distribution may not follow the conditional Gaussian distribution (e.g., mixture Gaussian, 2D distribution like S-curve, and image dataset), the forthcoming experiments will show that the proposed variance-scaled reverse sampling can effectively and flexibly traverse the DP tradeoff on general datasets using a single score network.

 \subsection{Variance-scaled reverse sampling process}
In this subsection,  we discuss how to perform variance-scaled reverse sampling by estimating the conditional score. With \emph{a single pre-trained score network}, we can traverse different DP tradeoffs for general inverse problems.

Recall that the proposed variance-scaled reverse sampling process is given by
    $$p_{\lambda}(\mathbf x_{0:T}|\mathbf y)=p_{\lambda}(\mathbf x_T|y)\prod_{k=0}^{T-1}p_{\lambda}(\mathbf x_{k}|\mathbf x_{k+1},\mathbf y),$$
 where for VP diffusion, we have the posteriors 
    $p_{\lambda}(\mathbf x_{k-1}|\mathbf x_{k},\mathbf y):=\mathcal{N}\big(\boldsymbol{\mu}_{k-1}(\mathbf x_{k}, \mathbf y), ~\lambda \mathbf C_{k-1}\big)$.
The sampling process in each step can be approximately expressed as 
\begin{align}
    \!\!\!\!\mathbf x_{k\!-\!1}\!=\!\frac{1}{\sqrt{\alpha}_k}(\mathbf x_k\!+\!(1\!-\!\alpha_k)\nabla_{\mathbf{x}_k}\log p(\mathbf x_k|\mathbf{y}))\!+\!\lambda\mathbf{C}_k\mathbf z, \label{Eq_sample_step_lambda}
\end{align}
where $\mathbf z\sim\mathcal{N}(0,\mathbf I)$. To draw a sample $\mathbf{x}_{k-1}$ according to \eqref{Eq_sample_step_lambda}, we need to compute the conditional score $\nabla_{\mathbf{x}_k}\log p(\mathbf x_k|\mathbf{y})$.  Specifically, the conditional scores can be decomposed as $\nabla_{\mathbf{x}_k}p(\mathbf{x}_k|\mathbf y) = \nabla_{\mathbf{x}_k}p(\mathbf{x}_k) + \nabla_{\mathbf{x}_k}p(\mathbf y|\mathbf{x}_k)$ by Bayes' rule, where the first term can be approximated by a score network $s_{\theta}(\mathbf{x}_k,k)$ \cite{DDPM,song2021scorebased}. To deal with the second term, Denoising Posterior Sampling (DPS) \cite{chung2023DPS} proposes to approximate $p(\mathbf y|\mathbf{x}_k)=\E_{p(\mathbf{x}_0|\mathbf{x}_k)}[p(\mathbf y|\mathbf{x}_0)]$ with $p(\mathbf y|\hat{\mathbf{x}}_0(\mathbf{x}_k))$, where 
\begin{align}
    \hat{\mathbf{x}}_0(\mathbf{x}_k) \!=\! \E_{p(\mathbf{x}_0|\mathbf{x}_k)}[X_0] \!=\! \frac{1}{\alpha_k}(\mathbf{x}_k\!+\!(1\!-\!\bar{\alpha}_k)\nabla_{\mathbf{x}_k}p(\mathbf{x}_k)) \label{Eq_x0_Tweedie}
\end{align}
$\!\!\!\text{by}$ Tweedie's formula. The approximation error here is upper-bounded by the Jensen gap \cite{Jesen_Gap}. Note that $p(\mathbf y|\hat{\mathbf{x}}_0(\mathbf{x}_k))$ is analytically tractable when the measurement distribution is given. Specifically, when the measurement nosie is $\mathcal{N}(0, \sigma_n^2\mathbf{I})$, the second term in the conditional score takes the form of
    $\nabla_{\mathbf{x}_k}\log p(\mathbf y|\mathbf{x}_k) = \frac{1}{\sigma_n^2} \nabla_{\mathbf{x}_k} ||\mathbf{y}-\mathcal{A}(\hat{\mathbf{x}}_0(\mathbf{x}_k))||^2_2.$
Thus, the conditional scores can be approximated as $\nabla_{\mathbf{x}_k}\log p(\mathbf{x}_k|\mathbf y) = s_{\theta}(\mathbf{x}_k,k) + \frac{1}{\sigma_n^2} \nabla_{\mathbf{x}_k} ||\mathbf{y}-\mathcal{A}(\hat{\mathbf{x}}_0(\mathbf{x}_k))||^2_2$.

In the following experiments, we adopt the DPS framework to approximate the conditional score and sample from \eqref{Eq_sample_step_lambda}, which is sketched as Algorithm \ref{alog_lam_DPS}.

\begin{algorithm}[H]
    \centering
    \caption{Variance-Scaled Reverse Sampling with DPS}\label{alog_lam_DPS}
    \begin{algorithmic}
        \Require $T$, $\mathbf{y}$, $\lambda$, $\{\tilde{\sigma_k}\}_{k=0}^{T}$, $\{\zeta_{k,\lambda}\}_{k=0}^T$
        \State $\mathbf{x}_T\sim\mathcal{N}(\mathbf{0},\mathbf{I})$
        \For{$k=T-1$ to $1$ do}
            \State $\hat{s} \gets s_{\theta}(\mathbf{x}_k, k)$
            \State $\hat{\mathbf{x}}_0 \gets \frac{1}{\sqrt{\bar{\alpha}_k}}(\mathbf x_k+(1-\bar{\alpha}_k)\hat{s})$
            \State $\hat{c}(\hat{\mathbf{x}}_0)=-\frac{\sqrt{\alpha_k}}{1-\alpha_k}\nabla_{\mathbf x_k}||y-\mathcal{A}(\hat{\mathbf x}_0(\mathbf x_k))||^2_2$
            \State $\mathbf{z}\sim\mathcal{N}(\mathbf{0},\mathbf{I})$
            \State $\mathbf{x}_{k-1} \gets \frac{1}{\sqrt{\alpha_k}}\big(\mathbf{x}_k + (1-\alpha_k)\big(\hat{s}+\zeta_{k,\lambda}\hat{c}(\hat{\mathbf{x}}_0)\big)\big) + \lambda\tilde{\sigma}_k\mathbf z$
        \EndFor\\
        \Return $\mathbf{x}_0$
    \end{algorithmic}
\end{algorithm}

In Algorithm \ref{alog_lam_DPS}, $\tilde{\sigma}_k$ is set to $\beta_k$ \cite{DDPM,Xue2024_scorebased_variational_infer}, and $\zeta_{k,\lambda}$ is a hyperparameter to control the weight of the conditional score, and may differ for different $\lambda$. The heuristic choices of $\zeta_{k,\lambda}$ and the rationale for fine-tuning have been included in Appendix \ref{Appendix_Exp_detail}.

 \begin{remark}
    In this paper, we adopt the conditional score approximation proposed in \cite{chung2023DPS} for illustration. To implement the variance-scaled reverse sampling process \eqref{Eq_sample_step_lambda}, one can employ any approach to directly estimate the conditional score \cite{song2023pseudoPGDM} or estimate the posterior mean as a whole \cite{Xue2024_scorebased_variational_infer} for general inverse problems. For example, Pseudoinverse-guided Diffusion Models ($\Pi$GDM) \cite{song2023pseudoPGDM} approximated the unconditional posteriors $p(\mathbf{x}_0|\mathbf{x}_k)$ with Gaussian, i.e., $p(\mathbf{x}_0|\mathbf{x}_k)\approx\mathcal{N}(\hat{\mathbf{x}}_0, r_k^2\mathbf{I})$, where $\hat{\mathbf{x}}_0$ is given by \eqref{Eq_x0_Tweedie}, and $r_k^2$ is the hyperparameter. Thus, the score term $\nabla_{\mathbf{x}_k}p(\mathbf y|\mathbf{x}_k)$ can be analytically expressed by the pseudoinverse of the measurement model. In \cite{Xue2024_scorebased_variational_infer}, the posterior mean is estimated as a whole. Specifically, we have $\boldsymbol{\mu}_{k-1}(\mathbf x_{k}, \mathbf y)=\mathbf U_{k-1}\mathbf x_{k}+\mathbf V_{k-1}\mathbb E_{p(\mathbf x_0|\mathbf y)}[X_0]$. Since the expression involves $\mathbb E_{p(\mathbf x_0|\mathbf y)}[X_0]$ and $\Cov_{p(\mathbf x_0|\mathbf y)}[X_0]$, it can not be computed directly. Instead, the authors approximate the joint conditional posterior with variational Gaussian distributions and apply natural gradient descent to perform the sampling. Note that the gradient calculation also involves the approximation of the conditional score $\nabla_{\mathbf{x}_k}\log p(\mathbf y|\mathbf{x}_k)$ as in DPS \cite{chung2023DPS}. 
\end{remark}

\subsection{An illustrative example}\label{sub-sec:mixture_gaussian}

\begin{figure}[t]
    \centering
    \includegraphics[width=0.5\textwidth]{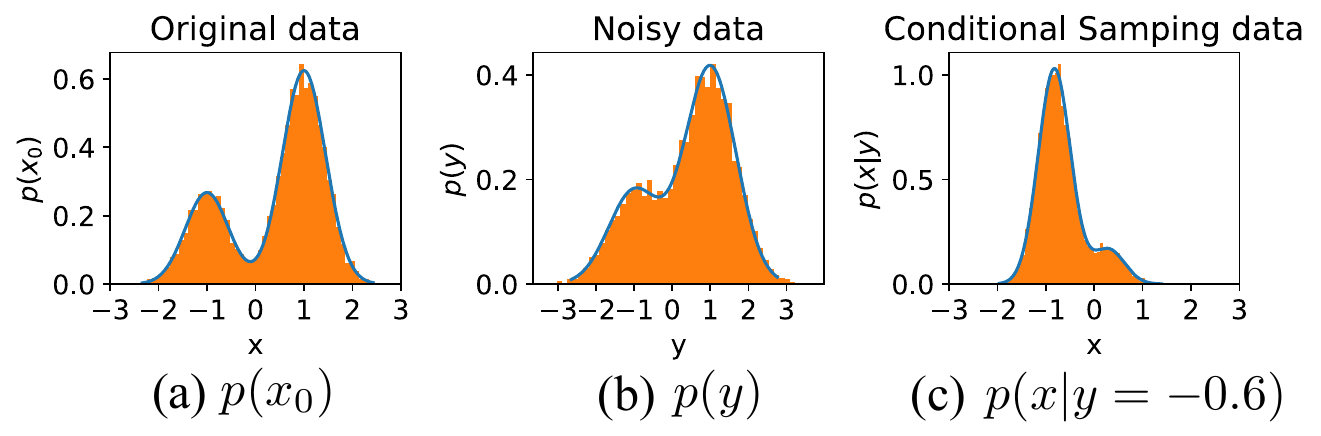}
    \caption{An example of mixture Gaussian, the noisy observation, and the conditional distribution given an observation $y=-0.6$}\label{Fig_mix_Gaussian_dist}
\end{figure}

Consider the mixture Gaussian distribution $X_0\sim p(x_0)$ with two components, where
    $p(x_0)=w_1\mathcal{N}(\mu_1, \sigma_1^2)+w_2\mathcal{N}(\mu_2, \sigma_2^2)$.
The noisy observation is obtained by $Y=aX_0+\sigma_0\epsilon$, where $\epsilon\sim\mathcal{N}(0,1)$, i.e., $p(y|x_0)=\mathcal{N}(ax_0, \sigma_0^2)$. The MMSE estimator and posterior distribution can be theoretically derived (see Appendix \ref{Appendix_MixtureGaussian} for details). Fig. \ref{Fig_mix_Gaussian_dist} shows an example of mixture Gaussian distribution and the conditional distribution given an observation $y=-0.6$.

\begin{figure*}[t]
    \centering
    \vspace{-0.5cm}
    \includegraphics[width=1\textwidth]{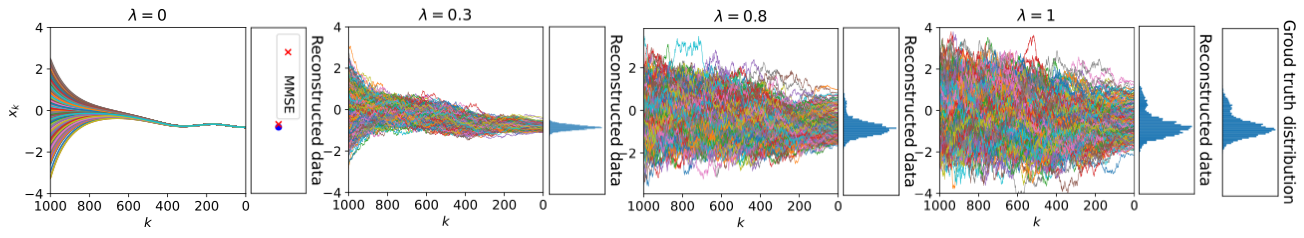}
    \caption{Trajectories $x_T\to \cdots x_k\to\cdots\to x_0$ of different reconstructions for $\lambda=0, 0.3, 0.8$ and $1$. The initial $x_T$ is $\mathcal{N}(0,1)$.}\label{Fig_traj_all}
    \vspace{-0.2cm}
\end{figure*}

Given an observation $y$ and a chosen $\lambda$, we iteratively perform the variance scaled reverse process \eqref{Eq_sample_step_lambda} for $k=T,\cdots,1$, where the variance is approximated by $\mathbf{C}_k=\beta_k$ as the conventional VP-diffusion. Starting from a random sample $x_T\sim\mathcal{N}(0,1)$, Fig. \ref{Fig_traj_all} shows the trajectories $x_T\to \cdots x_k\to\cdots x_0$ of multiple reconstructions for each $\lambda$. We can observe that for $\lambda=0$, given an initial $x_T$, the trajectories are deterministic and converge to the MMSE point. This phenomenon collaborates with results in \cite{Xue2024_scorebased_variational_infer}. When $\lambda$ increases, the generated trajectories follow the form of the posterior distribution and show more stochasticity. When $\lambda=1$, the reconstruction distribution matches the ground truth posterior. 

\begin{figure}[t]
    \centering
    \includegraphics[width=0.75\textwidth]{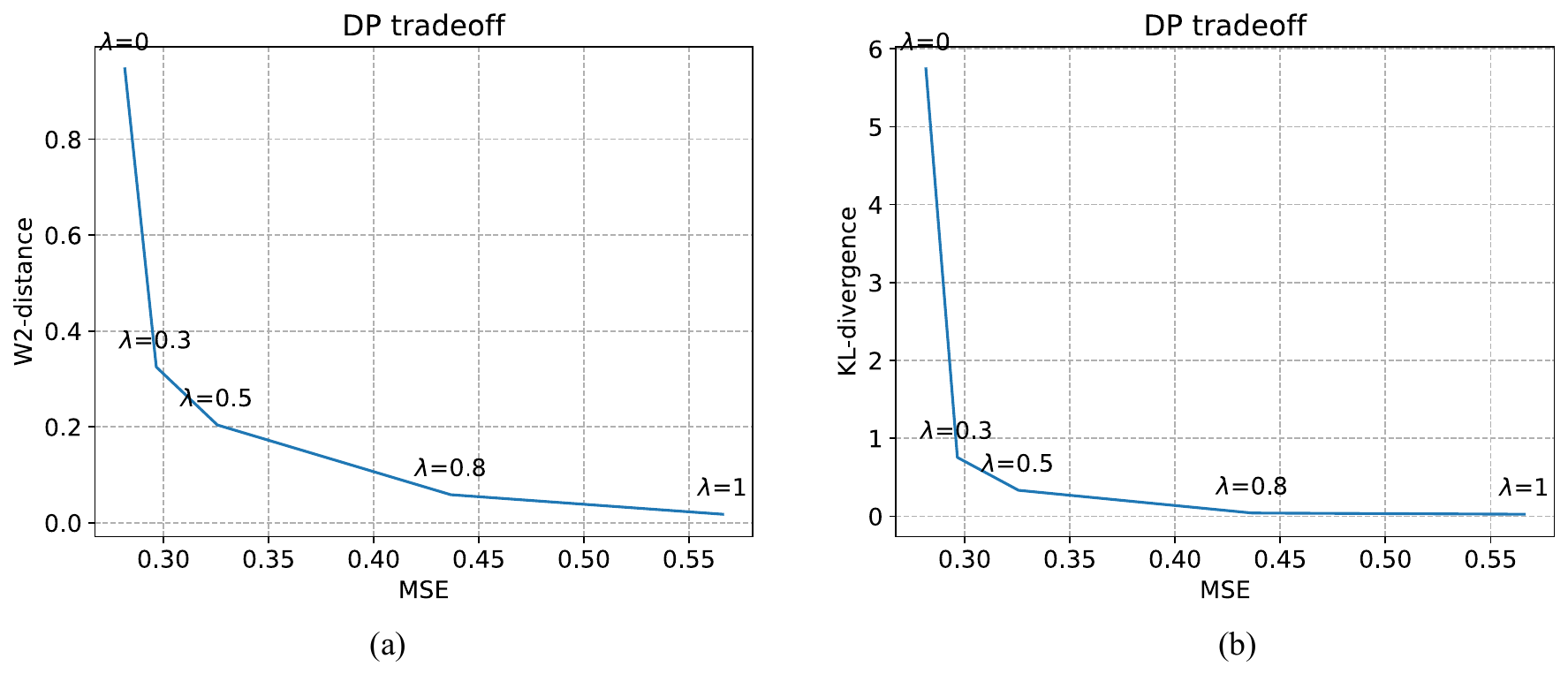}
    \vspace{-0.4cm}
    \caption{Distortion-perception tradeoff traversed by variance-scaled reverse sampling given different $\lambda$'s. (a) tradeoff between Wasserstein-2 distance and MSE; (b) tradeoff between KL-divergence and MSE.}\label{Fig_DP_mix_Gaussian}
    \vspace{-0.5cm}
\end{figure}

The resulting DP tradeoffs are shown in Fig. \ref{Fig_DP_mix_Gaussian}. It can be observed that as $\lambda$ increases, the divergence between the true posterior distribution and reconstruction distribution continuously decreases, with an increase of MSE value. The overall curve is convex, and both Wasserstein-2 and KL-divergence come to zero.  Meanwhile, the MSE achieved at $\lambda=1$ is approximately two times the MSE at $\lambda=0$ for both Wasserstein-2 and KL-divergence cases, coinciding with the result in Theorem \ref{thm_optimality_condi_Gaussian}. 

\section{Experimental Results}
In this section, we conduct a series of experiments on two-dimensional datasets and the FFHQ dataset \cite{FFHQ_dataset}. It is demonstrated that the variance-scaled reverse sampling shows great flexibility and effectiveness in traversing DP tradeoff compared to the GAN-based approach and other denoising methods. 

\subsection{Two-dimensional points example}\label{sub-sec:2DExpriments}
\begin{figure}[t]
    \centering
    \includegraphics[width=0.56\textwidth]{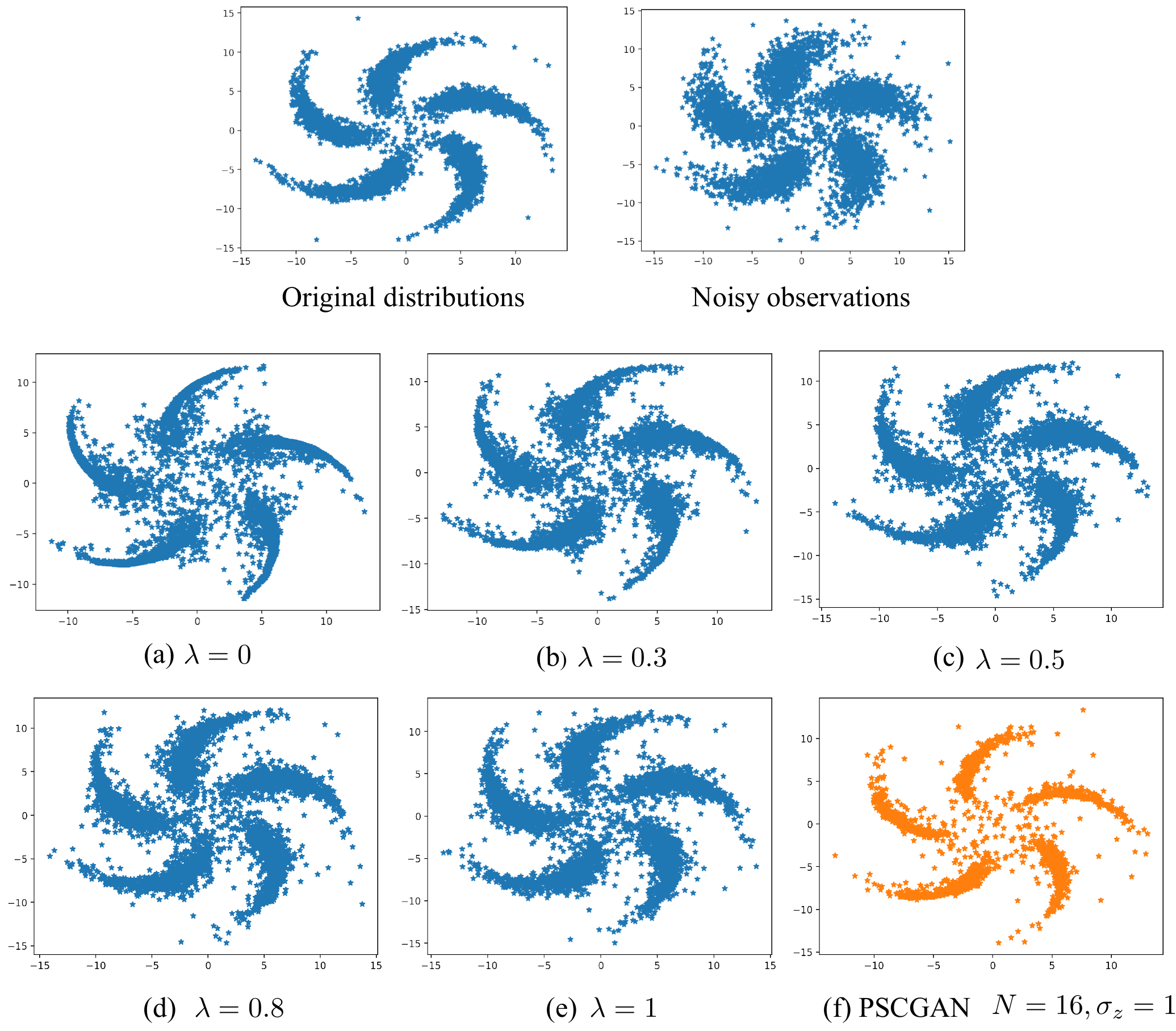}
    \caption{Experiments on a two-dimensional dataset. The first row illustrates the original distribution (left) of pinwheel and the noisy observation $Y$, where $Y=aX+N$ for $N\sim\mathcal{N}(0,\sigma_n^2\mathbf{I})$ (right). The second and third row shows the reconstructions:  (a)-(e) variance-scaled reverse diffusion process with different $\lambda$'s; (f) PSCGAN with $N=16,\sigma_z=1$.}\label{Fig_recon_pinwheel}
    \vspace{-0.3cm}
\end{figure}

First, we evaluate the validity and effectiveness of our approach on two-dimensional datasets, including pinwheel, S-curve, and moon-shape data distributions. Taking the pinwheel dataset as an example, the first row of Fig. \ref{Fig_recon_pinwheel} shows the distribution of pinwheel data points $X_0$ and the noisy observation $Y$, where $Y=aX+N$ for $N\sim\mathcal{N}(0,\sigma_n^2\mathbf{I})$.

For each dataset, we train a conventional score network to approximate $p(\mathbf{x}_0)$ on the original distribution. Then, we perform the variance-scaled reverse diffusion process \eqref{Eq_sample_step_lambda} on the noisy observation. As a comparison, we adopt a GAN-based DP-traversing approach PSCGAN \cite{PSCGAN}. This scheme introduces a penalty of the posterior expectation to the training of the conditional GAN. The authors proposed two strategies to navigate the DP tradeoff during the inference time: (1) PSCGAN-$N$: This method involves sampling $N$ instances from the generator and then averaging them. As $N$ increases, the averaged image is closer to the conditional expectation, leading to a smaller distortion but larger perception loss. (2) PSCGAN-$z$: It varies the standard deviation $\sigma_z$ of the noise injected into the generator to control the stochasticity. For $0\leq\sigma_z\leq 1$, a larger $\sigma_z$ results in a better perceptual quality and higher MSE. Note that the training of PSCGAN relies on noisy data points $Y$. Thus, the model needs to be retrained for different measurement scenarios and different levels of $\sigma_n^2$.

Fig. \ref{Fig_recon_pinwheel} shows the reconstruction of our variance-scaled reverse diffusion process for different $\lambda$'s, as well as the reconstruction of PSCGAN when setting $N=16, \sigma_z=1$. It can be observed that when $\lambda=0$, our sampling process leads to a more concentrated reconstruction, and gradually approaches the true distribution as $\lambda$ increases.

\begin{figure}[t]
    \centering
    \includegraphics[width=0.5\textwidth]{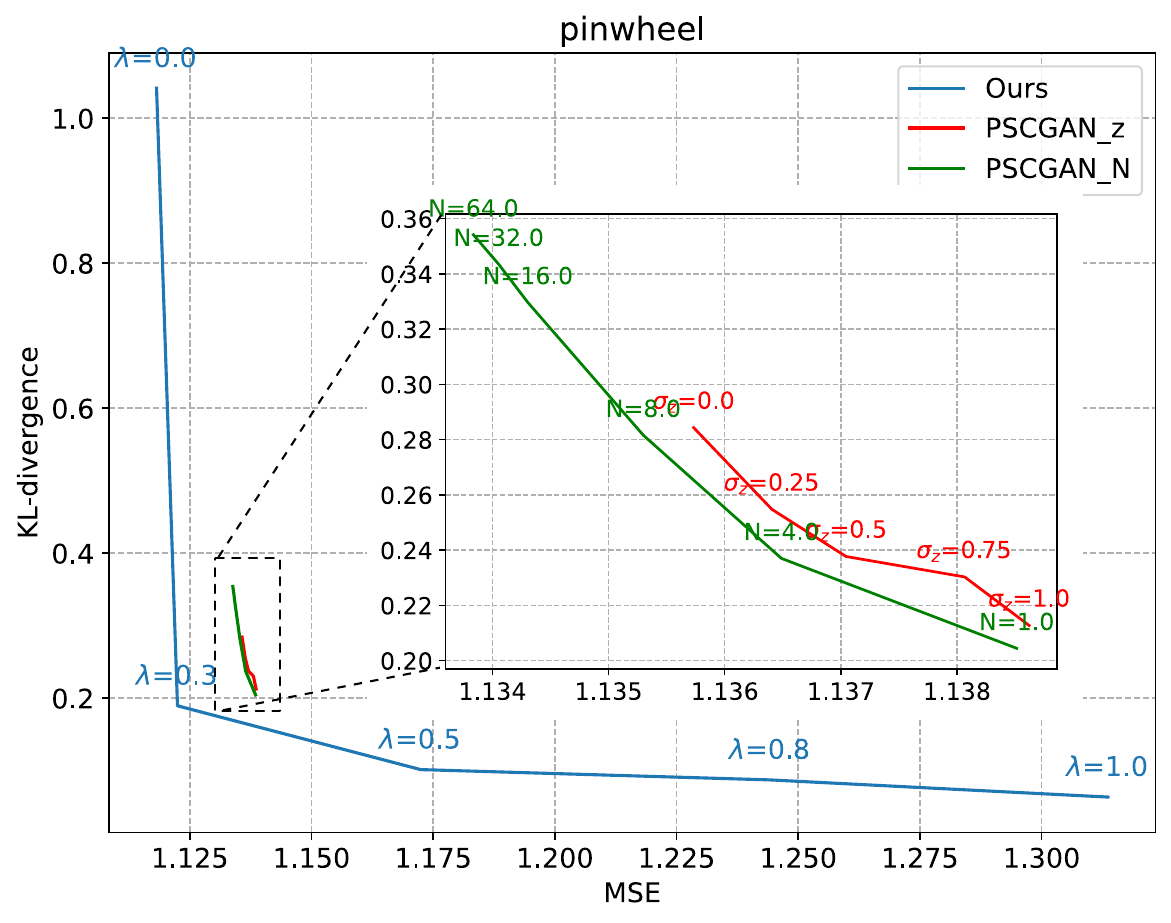}
    \caption{DP tradeoff on pinwheel dataset traversed by our variance-scaled reverse diffusion process and PSCGAN. }\label{Fig_DP_pinwheel}
    \vspace{-0.5cm}
\end{figure}

Fig. \ref{Fig_DP_pinwheel} compares the numerical DP tradeoffs achieved by our method versus PSCGAN. Our score-based method achieves a much larger range of tradeoffs compared to the GAN-based approach, demonstrating superior performance. Moreover, unlike PSCGAN, our method requires only a single score network to handle varying measurements and noise levels, offering greater flexibility. Additional experimental results on other distributions are provided in Appendix \ref{Appendix_more_exp}.

\subsection{FFHQ dataset}
In this subsection, we evaluate our method on the FFHQ $256\times 256$ dataset \cite{FFHQ_dataset}. For different measurement scenarios, including Gaussian blur and super-resolution, we use the same pre-trained score-based model from \cite{chung2023DPS}, which was trained from scratch using 49k training data for 1M steps. We consider measurement scenarios with noise levels more severe than those in \cite{chung2023DPS,Xue2024_scorebased_variational_infer,DiffPIR}. Specifically, for Gaussian blur, the images are blurred by a Gaussian blur kernel with size $61\times 61$ and a standard deviation of 3.0. 
For super-resolution, the images are downsampled by a factor of 8.

We compare our method against two benchmarks: (1) We compare the DP tradeoffs provided by our score-based method and PSCGAN \cite{PSCGAN} on the Gaussian deblur task. Note that PSCGAN is not designed for other measurement scenarios and requires retraining for different noise levels. (2) We evaluate the performance of DiffPIR \cite{DiffPIR} to see where it falls within the tradeoff spectrum for different inverse problems. DiffPIR is a plug-and-play image restoration method based on diffusion denoising implicit model (DDIM) \cite{song2021DDIM}. It has shown competitive performance on many denoising problems with fewer diffusion steps. 

\subsubsection{Gaussian deblur task}
\begin{figure*}[t]
    \centering
    \includegraphics[width=0.9\textwidth]{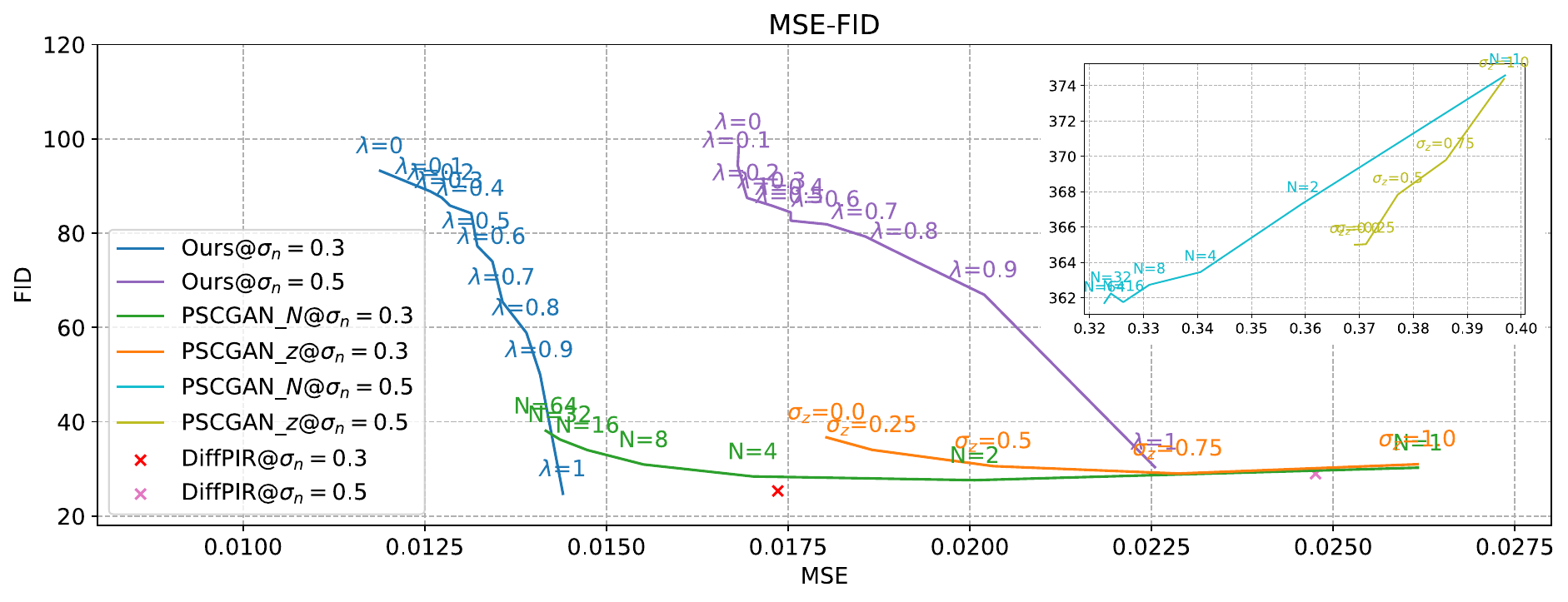}
    \caption{DP tradeoff on FFHQ dataset. Note that the PSCGAN model in this figure is trained on images contaminated with additive Gaussian noise of $\sigma_n=0.3$. It can be observed that when the noise level in inference time does match the trained model (e.g., when the noise level becomes $\sigma_n=0.5$ during the test time), PSCGAN fails to provide valid reconstructions. On the other hand, our variance-scaled reverse diffusion process and DiffPIR method rely on a pre-trained score network on the FFHQ dataset, which is not dedicated to any noisy measurement and is robust for different noise levels.}\label{Fig_DP_FFHQ_GS}
    \vspace{-0.3cm}
\end{figure*}

Fig. \ref{Fig_DP_FFHQ_GS} shows the DP tradeoff on the Gaussian deblur task. We test our variance-scaled reverse diffusion process, PSCGAN \cite{PSCGAN}, and DiffPIR \cite{DiffPIR} on two different additive noise levels $\sigma_n=0.3$ and $\sigma_n=0.5$. Note that our sampling method and DiffPIR rely on a pre-trained score network on the FFHQ dataset. For PSCGAN, we use the model trained on images with additive Gaussian noise of $\sigma_n=0.3$.

For noise level $\sigma_n=0.3$, PSCGAN shows suboptimal and more limited tradeoffs compared to our method. To obtain a reconstruction with similar MSE achieved in our $\lambda=0.9$ case, PSCGAN requires averaging 64 sampled images. Its distortion degrades rapidly when fewer images are averaged. Our variance-scaled reverse diffusion sampling achieves both better fidelity and a broader range of DP tradeoffs. Moreover, when the models are tested on the higher noise level of $\sigma_n=0.5$, the overall tradeoff shifts to the left. Our sampling method maintains effective tradeoff traversal. PSCGAN's performance (trained at $\sigma_n=0.3$) deteriorates significantly, showing much larger MSE and excessively high FID values.

\begin{figure*}[t]
    \centering
    \includegraphics[width=0.9\textwidth]{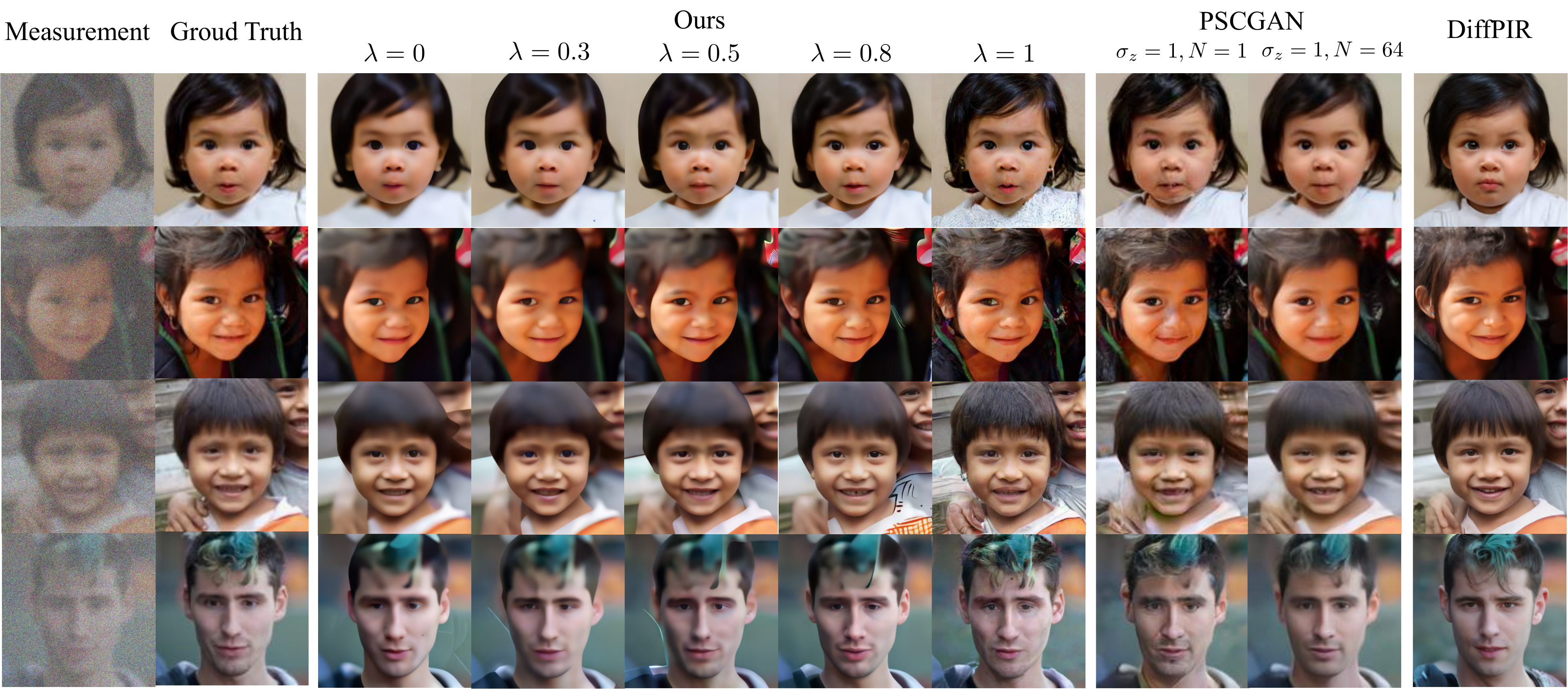}
    \caption{Samples from our sampling process, PSCGAN, and DiffPIR for Gaussian deblur task with noise level $\sigma_n=0.3$.}\label{Fig_recon_GS_03}
\end{figure*}

Significant visual improvements can be observed in reconstructed samples in Fig. \ref{Fig_recon_GS_03}. When $\lambda=0$, the restored images maintain high fidelity and faithfully reconstruct the original images, though appear relatively blurry. As $\lambda$ increases, the images become progressively sharper, revealing more fine details such as hair strands, eye corners, and background elements. When $\lambda=1$, while the overall reconstruction appears very natural, subtle details deviate from the original images. In comparison, both PSCGAN and DiffPIR have relatively good perceptual quality but compromise fidelity on crucial details. Notable deviations appear in features like eyes and mouth shapes, ultimately altering the facial expression and overall style of faces. We also report more metrics (PSNR and LPIPS) of Gaussian deblur task with additive noise level $\sigma_n = 0.3$ in Appendix \ref{Appendix_Exp_More_results}.

\subsubsection{Super-resolution task}
\begin{figure*}[t]
    \centering
    \includegraphics[width=0.96\textwidth]{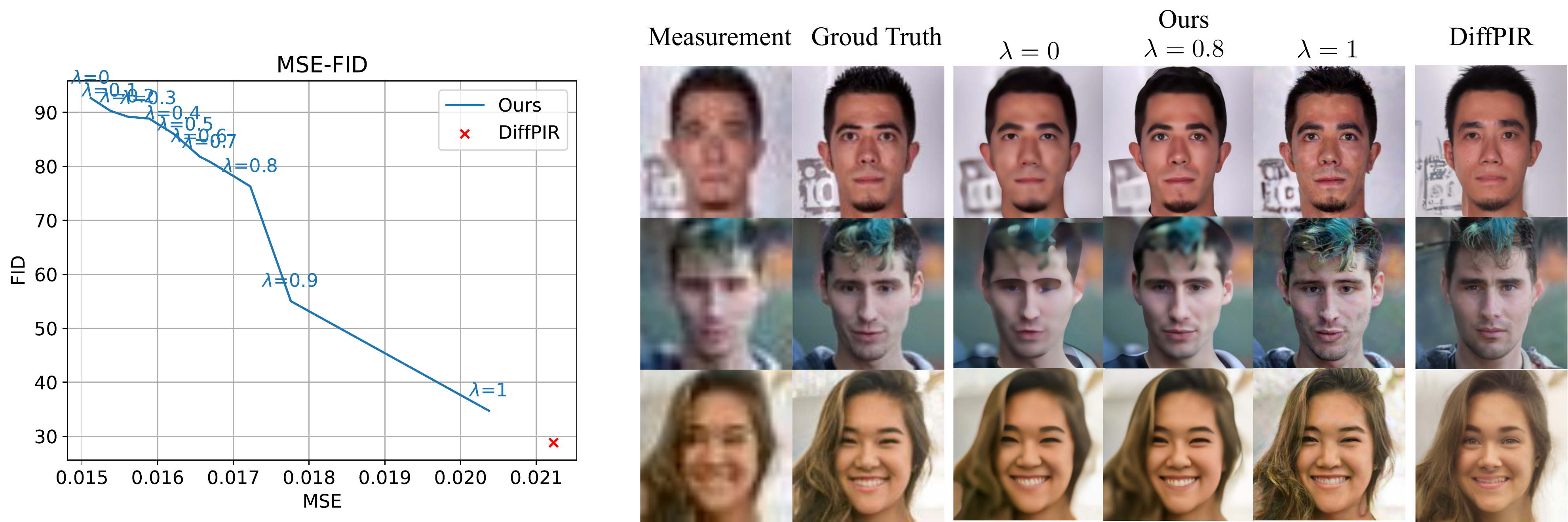}
    \caption{Samples from our sampling process, PSCGAN, and DiffPIR for super-resolution task with downsampling scale $\times 8$.}\label{Fig_DP_SR}
\end{figure*}

Similar phenomena are observed in the super-resolution (SR) task. By adjusting different $\lambda$'s, we can traverse a wide range of points on the DP plane, allowing flexible selection of the most suitable point for specific application scenarios. Fig. \ref{Fig_DP_SR} shows the numerical results and reconstruction samples given by our sampling process and DiffPIR. Note that DiffPIR performs close to our $\lambda=1$ case, and achieves a better FID score but larger distortion than our method. While DiffPIR demonstrates strong generative capabilities in both Gaussian deblur and super-resolution tasks with fewer diffusion steps \cite{DiffPIR}, we examine more challenging degradation scenarios ($\sigma_n = 0.3$ and 0.5 for deblurring, and $8\times$ downscaling for SR). Under these severe conditions, DiffPIR's reduced diffusion steps lead to fidelity issues, with generated details less faithful compared to our $\lambda=1$ case. Despite these differences, theoretical connections exist between DPS and DiffPIR \cite{DiffPIR}. We will consider the exploration of DDIM-style sampling for faster tradeoff traversal as future work.

In summary, using a single pre-trained score network, the proposed sampling method can effectively traverse a more complete DP tradeoff and achieve better MSE than the benchmarks. Experimental details and more results are included in Appendix \ref{Appendix_more_exp}.

\section{Conclusions}\label{SecConclusion}
In this paper, we proposed a variance-scaled reverse diffusion process to traverse the DP tradeoff for general inverse problems using a single score-based model. By tuning a parameter that controls the scale of the reverse variance, we can navigate from the MMSE point to the whole posterior distribution, generating a complete DP curve. We proved that the proposed reverse sampling process serves as an optimal solution to the conditional DP tradeoff for multivariate Gaussian distribution. Meanwhile, we conducted experiments on the mixture Gaussian example, two-dimensional datasets, and the FFHQ dataset. Our results show that using a single pre-trained score network, we have achieved a more complete empirical DP tradeoff than the GAN-based method and other inverse problem solvers, demonstrating the effectiveness and flexibility of the proposed framework.

{\appendices


\section{Approximation of Reverse Posterior Distribution}\label{Appendix_approx_reverse_condi}
In this section, we will first include the deviation of posterior mean and variance in \cite{Xue2024_scorebased_variational_infer} for self-contained. We will use a modified proof to reveal the relationship between the posterior mean and conditional score.

With the Bayes' rule, we have for VP-diffusion
\begin{align}
    p(\mathbf x_k|\mathbf x_{k+1}, \mathbf y)&=\frac{p(\mathbf x_{k+1}|\mathbf x_k)p(\mathbf x_k|\mathbf y)}{p(\mathbf x_{k+1}|\mathbf y)}\notag\\
    &\propto \exp\Big(-\frac{1}{2\beta_{k+1}}||\mathbf x_{k+1}-(1-\frac{1}{2}\beta_{k+1})\mathbf x_k||^2+\log p(\mathbf x_k|\mathbf y)-\log p(\mathbf x_{k+1}|\mathbf y)\Big).\label{exp}
\end{align}
We can approximate $\log p(\mathbf x_{k}|\mathbf y)$ by Taylor's expansion on point $\mathbf x_{k+1}$. When $T\to\infty$,
\begin{align*}
    \log p(\mathbf x_{k}|\mathbf y)&\approx \log p(\mathbf x_{k+1}|\mathbf y)+(\mathbf x_{k}-\mathbf x_{k+1})^{\top}\nabla_{\mathbf x_{k+1}}\log p(\mathbf x_{k+1}|\mathbf y)+\mathcal{O}(\|\mathbf x_{k}-\mathbf x_{k+1}\|^2).
\end{align*}
Then, \eqref{exp} can be written as 
\begin{align}
    p(\mathbf x_k|\mathbf x_{k+1},\mathbf y) &\propto \exp\bigg(-\frac{1}{2\beta_{k+1}}||\mathbf x_{k+1}-(1-\frac{1}{2}\beta_{k+1})\mathbf x_k||^2+(\mathbf x_{k}-\mathbf x_{k+1})^{T}\nabla_{\mathbf x_{k+1}}\log p(\mathbf x_{k+1}|\mathbf y)\bigg)\notag\\
    &=\exp\bigg(-\frac{1}{2 \beta_{k+1}}\Big(\mathbf x_{k+1}^{\top} \mathbf x_{k+1}-2(1-\frac{1}{2} \beta_{k+1}) \mathbf x_{k+1}^{\top} \mathbf x_k+(1-\frac{1}{2} \beta_{k+1})^2 \mathbf x_k^{\top} \mathbf x_k\Big)\notag\\
    &~~~~~~~~~~~~~~~~~~~~~~~~~~~~~~~~~~~~~~~~~~ + \mathbf x_k^{\top} \nabla_{\mathbf x_{k+1}} \log p(\mathbf x_{k+1} \mid \mathbf y)-\mathbf x_{k+1}^{\top} \nabla_{\mathbf x_{k+1}} \log p(\mathbf x_{k+1}|\mathbf y)\bigg)\notag \\
    &\propto\exp\bigg(-\frac{1-\beta_{k+1}}{2 \beta_{k+1}} \mathbf x_k^{\top}\mathbf x_k+\Big(\frac{1-\frac{1}{2} \beta_{k+1}}{\beta_{k+1}} \mathbf x_{k+1}+\nabla_{\mathbf x_{k+1}} \log p(\mathbf x_{k+1}|\mathbf y)\Big)^{\top} \mathbf x_{k}\bigg)\notag\\  
    &\propto \exp\bigg(-\frac{1-\beta_{k+1}}{2\beta_{k+1}}\Big(\mathbf x_k^{\top}\mathbf x_k-2\big(\frac{1-\frac{1}{2}\beta_{k+1}}{1-\beta_{k+1}}\mathbf x_{k+1}-\frac{\beta_{k+1}}{1-\beta_{k+1}}\nabla_{\mathbf x_{k+1}} \log p(\mathbf x_{k+1}|\mathbf y)\big)^{\top}\mathbf x_k\Big)\bigg)\notag\\
    &\propto \exp\bigg(-\frac{1-\beta_{k+1}}{2\beta_{k+1}}\Big\|\mathbf x_{k}-\frac{1}{\sqrt{\alpha_{k+1}}}\Big(\mathbf x_{k+1}+(1-\alpha_{k+1})\nabla_{\mathbf x_{k+1}}\log p(\mathbf x_{k+1}|\mathbf y)\Big)\Big\|^2\bigg),\label{condi_posterior_score}
\end{align}
where the last step utilizes the equivalent infinitesimal $\sqrt{\alpha_k}=\sqrt{1-\beta_{k}}= 1-\frac{1}{2}\beta_{k}$ when $T\to \infty$.
From \eqref{condi_posterior_score}, we can see that $p(\mathbf x_k|\mathbf x_{k+1},\mathbf y)$ has mean 
\begin{align}
    \boldsymbol\mu_k(\mathbf{x}_{k+1}, \mathbf y)\triangleq \frac{1}{\sqrt{\alpha_{k+1}}}\Big(\mathbf x_{k+1}+(1-\alpha_{k+1})\nabla_{\mathbf x_{k+1}}\log p(\mathbf x_{k+1}|\mathbf y)\Big).\label{mean_score}
\end{align}

For VP diffusion, the expectation and covariance of $p(\mathbf x_k|\mathbf y)$ can be computed as \cite{Xue2024_scorebased_variational_infer}
\begin{align}
    \boldsymbol\mu_k = \mathbb{E}_{p\left(\mathbf{x}_k \mid \mathbf{y}\right)}\left[X_k\right] & =\int \mathbf{x}_k p\left(\mathbf{x}_k \mid \mathbf{y}\right) d \mathbf{x}_k\notag \\
    & =\int \mathbf{x}_k \int p\left(\mathbf{x}_k \mid \mathbf{x}_0\right) p\left(\mathbf{x}_0 \mid \mathbf{y}\right) d \mathbf{x}_0 d \mathbf{x}_k \notag\\
    & =\iint \mathbf{x}_k p\left(\mathbf{x}_k \mid \mathbf{x}_0\right) d \mathbf{x}_k p\left(\mathbf{x}_0 \mid \mathbf{y}\right) d \mathbf{x}_0 \notag\\
    & =\int \sqrt{\bar{\alpha}_k} \mathbf{x}_0 p\left(\mathbf{x}_0 \mid \mathbf{y}\right) d \mathbf{x}_0=\sqrt{\bar{\alpha}_k} \mathbb{E}_{p\left(\mathbf{x}_0 \mid \mathbf{y}\right)}\left[X_0\right]\label{mu_k}\\
    \mathbf \Sigma_k=\operatorname{Cov}_{p\left(\mathbf{x}_k \mid \mathbf{y}\right)}\left[X_k\right]&=\int \mathbf{x}_k \mathbf{x}_k^\top p\left(\mathbf{x}_k \mid \mathbf{y}\right) d \mathbf{x}_k-\bar{\alpha}_k \mathbb{E}_{p\left(\mathbf{x}_0 \mid \mathbf{y}\right)}\left[X_0\right] \mathbb{E}_{p\left(\mathbf{x}_0 \mid \mathbf{y}\right)}\left[X_0\right]^\top\notag\\
    &=\int \mathbf{x}_k \mathbf{x}_k^\top \int p\left(\mathbf{x}_k \mid \mathbf{x}_0\right) p\left(\mathbf{x}_0 \mid \mathbf{y}\right) d \mathbf{x}_0 d \mathbf{x}_k-\bar{\alpha}_k \mathbb{E}_{p\left(\mathbf{x}_0 \mid \mathbf{y}\right)}\left[X_0\right] \mathbb{E}_{p\left(\mathbf{x}_0 \mid \mathbf{y}\right)}\left[X_0\right]^\top\notag\\
    &=\iint \mathbf{x}_k \mathbf{x}_k^\top p\left(\mathbf{x}_k \mid \mathbf{x}_0\right) d \mathbf{x}_k p\left(\mathbf{x}_0 \mid \mathbf{y}\right) d \mathbf{x}_0-\bar{\alpha}_k \mathbb{E}_{p\left(\mathbf{x}_0 \mid \mathbf{y}\right)}\left[X_0\right] \mathbb{E}_{p\left(\mathbf{x}_0 \mid \mathbf{y}\right)}\left[X_0\right]^\top\notag\\
    &=\int\left(\bar{\alpha}_k \mathbf{x}_0 \mathbf{x}_0^\top+\left(1-\bar{\alpha}_k\right) \mathbf{I}\right) p\left(\mathbf{x}_0 \mid \mathbf{y}\right) d \mathbf{x}_0-\bar{\alpha}_k \mathbb{E}_{p\left(\mathbf{x}_0 \mid \mathbf{y}\right)}\left[X_0\right] \mathbb{E}_{p\left(\mathbf{x}_0 \mid \mathbf{y}\right)}\left[X_0\right]^\top\notag\\
    &=\left(1-\bar{\alpha}_k\right) \mathbf{I}+\bar{\alpha}_k\left(\operatorname{Cov}_{p\left(\mathbf{x}_0 \mid \mathbf{y}\right)}\left[X_0\right]+\mathbb{E}_{p\left(\mathbf{x}_0 \mid \mathbf{y}\right)}\left[X_0\right] \mathbb{E}_{p\left(\mathbf{x}_0 \mid \mathbf{y}\right)}\left[X_0\right]^\top\right)-\bar{\alpha}_k \mathbb{E}_{p\left(\mathbf{x}_0 \mid \mathbf{y}\right)}\left[X_0\right] \mathbb{E}_{p\left(\mathbf{x}_0 \mid \mathbf{y}\right)}\left[X_0\right]^\top\notag\\
    &=\left(1-\bar{\alpha}_k\right) \mathbf{I}+\bar{\alpha}_k \operatorname{Cov}_{p\left(\mathbf{x}_0 \mid \mathbf{y}\right)}\left[X_0\right].\label{Sigma_k}
\end{align}

Suppose that $p(\mathbf x_k|\mathbf y)=\mathcal{N}(\boldsymbol{\mu}_k,\mathbf \Sigma_k)$, i.e., $\nabla_{x_k}\log p(\mathbf x_k|\mathbf y)=-\mathbf \Sigma_k^{-1}(\mathbf x_k-\boldsymbol{\mu}_k)$. We can obtain an approximation of the posterior mean $\boldsymbol{\mu}_{k-1}(\mathbf{x}_{k}, \mathbf y)$ from \eqref{mean_score}:
\begin{align*}
    \boldsymbol{\mu}_{k-1}(\mathbf{x}_{k}, \mathbf y)& = \frac{1}{\sqrt{\alpha}_k}(\mathbf x_k+(1-\alpha_k)\nabla_{\mathbf x_k}\log p(\mathbf x_k|\mathbf y))\\
    &=\frac{1}{\sqrt{\alpha}_k}(\mathbf x_k-(1-\mathbf \alpha_k)\mathbf\Sigma_k^{-1}(\mathbf x_k-\boldsymbol{\mu}_k))\\
    &=\frac{1}{\sqrt{\alpha}_k}(\mathbf I-(1-\alpha_k)\mathbf \Sigma_{k}^{-1})\mathbf x_k + \frac{1}{\sqrt{\alpha}_k}(1-\alpha_k)\mathbf \Sigma_k^{-1}\boldsymbol{\mu}_k\\
    &=\frac{1}{\sqrt{\alpha}_k}\big((1-\bar{\alpha}_k)\mathbf{I} + \bar{\alpha}_k\text{Cov}_{p(\mathbf x_0|\mathbf y)}[X_0]-(1-\alpha_k)\mathbf I\big)\big((1-\bar{\alpha}_k)\mathbf{I} + \bar{\alpha}_k\text{Cov}_{p(\mathbf x_0|\mathbf y)}[X_0]\big)^{-1}\mathbf x_k\\
    &~~~~~~~~~~ + \frac{1}{\sqrt{\alpha}_k}(1-\alpha_k)\mathbf \Sigma_k^{-1}\sqrt{\bar{\alpha}_k}\mathbb{E}_{p(\mathbf x_0|\mathbf y)}[X_0]\\
    &=\frac{1}{\sqrt{\alpha}_k}\alpha_k\big((1-\bar{\alpha}_{k-1})\mathbf{I} + \bar{\alpha}_{k-1}\text{Cov}_{p(\mathbf x_0|\mathbf y)}[X_0]\big)\big((1-\bar{\alpha}_k)\mathbf{I} + \bar{\alpha}_k\text{Cov}_{p(\mathbf x_0|\mathbf y)}[X_0]\big)^{-1}\mathbf x_k\\
    &~~~~~~~~~~  + (1-\alpha_k)\mathbf \Sigma_k^{-1}\sqrt{\bar{\alpha}_{k-1}}\mathbb{E}_{p(\mathbf x_0|\mathbf y)}[X_0]\\
    &=\big((1-\bar{\alpha}_{k-1})\mathbf{I} + \bar{\alpha}_{k-1}\text{Cov}_{p(\mathbf x_0|\mathbf y)}[X_0]\big)\big((1-\bar{\alpha}_k)\mathbf{I} + \bar{\alpha}_k\text{Cov}_{p(\mathbf x_0|\mathbf y)}[X_0]\big)^{-1}\sqrt{\alpha_k}\mathbf x_k\\
    &~~~~~~~~~~  +(1-\alpha_k)\mathbf \Sigma_k^{-1}\sqrt{\bar{\alpha}_{k-1}}\mathbb{E}_{p(\mathbf x_0|\mathbf y)}[X_0],
\end{align*}
which is the mean derived in \cite{Xue2024_scorebased_variational_infer}.


We can also utilize the Gaussian assumption to compute the posterior distribution $p(\mathbf x_k|\mathbf x_{k+1},\mathbf y)$ with the following lemma.
\begin{lem}{\cite[Section 2.3.3]{PRML}}\label{Gaussian_lem}
    Given a marginal Gaussian distribution for $X$ and a conditional Gaussian distribution for $Y$ given $X$ in the form
    \begin{align*}
    p(\mathbf{x}) & =\mathcal{N}\left(\mathbf{x} \mid \boldsymbol{\mu}, \boldsymbol{\Lambda}^{-1}\right), \\
    p(\mathbf{y} \mid \mathbf{x}) & =\mathcal{N}\left(\mathbf{y} \mid \mathbf{A} \mathbf{x}+\mathbf{b}, \mathbf{L}^{-1}\right),
    \end{align*}
    the marginal distribution of $Y$ and the conditional distribution of $X$ given $Y$ are given by
    \begin{align*}
    p(\mathbf{y}) & =\mathcal{N}\left(\mathbf{y} \mid \mathbf{A} \boldsymbol{\mu}+\mathbf{b}, \mathbf{L}^{-1}+\mathbf{A} \mathbf{\Lambda}^{-1} \mathbf{A}^{\top}\right) \\
    p(\mathbf{x} \mid \mathbf{y}) & =\mathcal{N}\left(\mathbf{x} \mid \boldsymbol{\Sigma}\left\{\mathbf{A}^{\top} \mathbf{L}(\mathbf{y}-\mathbf{b})+\boldsymbol{\Lambda} \boldsymbol{\mu}\right\}, \boldsymbol{\Sigma}\right),
    \end{align*}
    where
    $$
    \boldsymbol{\Sigma}=\left(\boldsymbol{\Lambda}+\mathbf{A}^{\top} \mathbf{L} \mathbf{A}\right)^{-1}.
    $$
\end{lem}

Suppose that $p(\mathbf x_k|\mathbf y)=\mathcal{N}(\boldsymbol{\mu}_k,\mathbf \Sigma_k)$, i.e., $\nabla_{\mathbf x_k}\log p(\mathbf x_k|\mathbf y)=-\mathbf \Sigma_k^{-1}(\mathbf x_k-\boldsymbol{\mu}_k)$. Together with $p(\mathbf x_{k+1}|\mathbf x_k)=\mathcal{N}(\sqrt{1\!\!-\!\!\beta_{k+1}}\mathbf x_k,\beta_{k+1}\mathbf I)$, we can directly obtain that $p(\mathbf x_{k}|\mathbf x_{k+1},\mathbf y)$ is a Gaussian with mean $\boldsymbol{\mu}_k(\mathbf x_{k+1}, \mathbf y)=\mathbf U_{k}\mathbf x_{k+1}+\mathbf V_{k}\E_{p(\mathbf x_0|\mathbf y)}[X_0]$ and variance $\mathbf C_k$ where

\begin{align}
    \mathbf U_{k} &:= \sqrt{\alpha_{k+1}}\big((1-\bar{\alpha}_{k})\mathbf I+\bar{\alpha}_{k}\text{Cov}_{p(\mathbf x_0|\mathbf y)}[X_0]\big)\big((1-\bar{\alpha}_{k+1})\mathbf I+\bar{\alpha}_{k+1}\text{Cov}_{p(\mathbf x_0|\mathbf y)}[X_0]\big)^{-1} \notag\\
    &= (1-\frac{1}{2}\beta_{k+1})\mathbf \Sigma_k\big((1-\beta_{k+1})\mathbf \Sigma_k+\beta_{k+1}\mathbf I\big)^{-1} \label{Eq_U_k}
    \\
    \mathbf V_{k} &:=  \sqrt{\bar{\alpha}_k}(1-\alpha_{k+1})\big((1-\bar{\alpha}_{k+1})\mathbf I+\bar{\alpha}_{k+1}\text{Cov}_{p(\mathbf x_0|\mathbf y)}[X_0]\big)^{-1} \notag \\
    & = \beta_{k+1}\sqrt{\bar{\alpha}_k}\big((1-\beta_{k+1})\mathbf \Sigma_k+\beta_{k+1}\mathbf I\big)^{-1} \label{Eq_V_k}
    \\
    \mathbf C_k &:= \frac{\beta_{k+1}}{1-\beta_{k+1}}\big((1-\bar{\alpha}_k)\mathbf I+\bar{\alpha}_k\text{Cov}_{p(\mathbf x_0|\mathbf y)}[X_0]\big)\Big(\big(\frac{\beta_{k+1}}{1-\beta_{k+1}}+1-\bar{\alpha}_k\big)\mathbf I+\bar{\alpha}_k\text{Cov}_{p(\mathbf x_0|\mathbf y)}[X_0]\Big)^{-1}\notag\\
    &= \beta_{k+1}\mathbf \Sigma_k\big((1-\beta_{k+1})\mathbf \Sigma_k+\beta_{k+1}\mathbf I\big)^{-1}.\label{Eq_C_k}
\end{align}
For each parameter $\mathbf U_{k}$, $\mathbf V_{k}$, and $\mathbf C_k$, the first expression is used in \cite{Xue2024_scorebased_variational_infer}. The second expression is equivalent when considering the equivalent infinitesimal $\sqrt{1-\beta_{k}} = 1-\frac{1}{2}\beta_{k}$ as $T\to \infty$, and will be used in the following proofs for convenience.

\section{Proof of Theorem \ref{thm_reverse_mean_var}}\label{Appendix_Proof_of_scaled_mean_var}
Since $p_{\lambda}(\mathbf x_{T-1}|\mathbf x_T,\mathbf y)=\mathcal{N}\Big(\mathbf U_{T-1}\mathbf x_{T}+\mathbf V_{T-1}\mathbb E_{p(\mathbf x_0|\mathbf y)}[X_0], \lambda \mathbf C_{T-1}\Big)$ and $p_{\lambda}(\mathbf x_T|\mathbf y)=\mathcal{N}(0,\mathbf I)$, from Lemma \ref{Gaussian_lem} we have that 
    \begin{align*}
        p_{\lambda}(\mathbf x_{T-1}|\mathbf y)=\mathcal{N}\Big(\mathbf V_{T-1}\mathbb E_{p(\mathbf x_0|\mathbf y)}[X_0],~ \lambda \mathbf C_{T-1}+\mathbf U_{T-1}\mathbf U_{T-1}^\top\Big).
    \end{align*}
    By simplifying the mean and variance, we have that 
    \begin{align}
        \boldsymbol{\mu}^{\lambda}_{T-1}&=\mathbf V_{T-1}\mathbb E_{p(\mathbf x_0|\mathbf y)}[X_0],\notag\\ 
        \mathbf \Sigma^{\lambda}_{T-1}&=\lambda \mathbf C_{T-1}+\mathbf U_{T-1}^\top\mathbf U_{T-1}\notag\\
        &=\lambda\beta_{T}\mathbf \Sigma_{T-1}\big((1\!\!-\!\!\beta_{T})\mathbf \Sigma_{T-1}+\beta_{T}\mathbf I\big)^{-1} + (1\!\!-\!\!\beta_{T})\mathbf \Sigma_{T-1}\big((1\!\!-\!\!\beta_{T})\mathbf \Sigma_{T-1}\!+\!\beta_{T}\mathbf I\big)^{-1} \big((1\!\!-\!\!\beta_{T})\mathbf \Sigma_{T-1}+\beta_{T}\mathbf I\big)^{-\top}\mathbf\Sigma_{T-1}^{\top} \notag\\
        &=\lambda\beta_{T}\mathbf \Sigma_{T-1}\big((1-\bar{\alpha}_{T})\mathbf I+\bar{\alpha}_{T}\Cov_{p(\mathbf x_0|\mathbf y)}[X_0]\big)^{-1}
        + \mathbf \Sigma_{T-1}\big((1-\bar{\alpha}_{T})\mathbf I+\bar{\alpha}_{T}\Cov_{p(\mathbf x_0|\mathbf y)}[X_0]\big)^{-1}\notag\\
        &~~~~~~~~~~~~~~~~~\cdot\big((1-\bar{\alpha}_{T})\mathbf I+\bar{\alpha}_{T}\Cov_{p(\mathbf x_0|\mathbf y)}[X_0]\big)^{-\top} \big((\alpha_T-\bar{\alpha}_T)\mathbf I + \bar{\alpha}_T\Cov_{p(\mathbf x_0|\mathbf y)}[X_0]\big)^{\top} \label{mid_sigma}\\
        &=(\lambda\beta_T+\alpha_T)\mathbf \Sigma_{T-1}, ~~~\text{as } \bar{\alpha}_T\to 0,\notag
    \end{align}
    where \eqref{mid_sigma} follows from $\mathbf{\Sigma}_{T-1}=(1-\bar{\alpha}_{T-1})\mathbf I+\bar{\alpha}_{T-1}\Cov_{p(\mathbf x_0|\mathbf y)}[\mathbf x_0]$.

    Then at time $T-2$, since $p_{\lambda}(\mathbf x_{T-2}|\mathbf x_{T-1},\mathbf y)=\mathcal{N}\Big(\mathbf U_{T-2}\mathbf x_{T-1}+\mathbf V_{T-2}\E_{p(\mathbf x_0|\mathbf y)}[X_0], \lambda \mathbf C_{T-2}\Big)$, we have that
    \begin{align*}
        p_{\lambda}(\mathbf x_{T-2}|\mathbf y)&=\mathcal{N}\Big((\mathbf U_{T-2}\mathbf V_{T-1}+\mathbf V_{T-2})\E_{p(\mathbf x_0|\mathbf y)}[X_0], ~ \lambda \mathbf C_{T-2}+\mathbf U_{T-2}(\lambda \mathbf C_{T-1}+\mathbf U_{T-1}\mathbf U_{T-1}^\top)\mathbf U_{T-2}^\top\Big)\\
        &=\mathcal{N}\Big((\mathbf U_{T-2}\mathbf V_{T-1}+\mathbf V_{T-2})\E_{p(\mathbf x_0|\mathbf y)}[X_0], ~ \lambda \mathbf C_{T-2}+\mathbf U_{T-2} (\lambda\beta_T+\alpha_T)\mathbf\Sigma_{T-1} \mathbf U_{T-2}^\top\Big),
    \end{align*}
    and we can further simplify the mean and variance as

    \begin{small}
    \begin{align*}
        \boldsymbol{\mu}^{\lambda}_{T-2} &= (\mathbf U_{T-2}\mathbf V_{T-1}+\mathbf V_{T-2})\E_{p(\mathbf x_0|\mathbf y)}[X_0]\\
        &=\Big(\sqrt{\alpha_{T-1}}\sqrt{\bar{\alpha}_{T-1}}(1-\alpha_{T})\big((1-\bar{\alpha}_{T-2})\mathbf I+\bar{\alpha}_{T-2}\text{Cov}_{p(\mathbf x_0|\mathbf y)}[X_0]\big)\big((1-\bar{\alpha}_{T})\mathbf I+\bar{\alpha}_{T}\text{Cov}_{p(\mathbf x_0|\mathbf y)}[X_0]\big)^{-1}\\
        &~~~~~~~~~~~~~~~~~~~~~~~~~~~~~~~~~~~~~~~~~~~~~~~~  
        +\sqrt{\bar{\alpha}_{T-2}}(1-\alpha_{T-1})\mathbf I\Big)\cdot\big((1-\bar{\alpha}_{T-1})\mathbf I+\bar{\alpha}_{T-1}\text{Cov}_{p(\mathbf x_0|\mathbf y)}[X_0]\big)^{-1}\E_{p(\mathbf x_0|\mathbf y)}[X_0] \\
        &=(1-\alpha_{T-1}\alpha_{T})\sqrt{\bar{\alpha}_{T-2}}\big((1-\bar{\alpha}_{T})\mathbf I+\bar{\alpha}_{T}\text{Cov}_{p(\mathbf x_0|\mathbf y)}[X_0]\big)^{-1}\E_{p(\mathbf x_0|\mathbf y)}[X_0] \\
        &=(1-\alpha_{T-1}\alpha_{T})\sqrt{\bar{\alpha}_{T-2}}\E_{p(\mathbf x_0|\mathbf y)}[X_0] ~~~\text{as } \bar{\alpha}_T\to 0,\\ 
        \mathbf \Sigma^{\lambda}_{T-2} &= \lambda \mathbf C_{T-2}+\mathbf U_{T-2} (\lambda\beta_T+\alpha_T\mathbf \Sigma_{T-1}) \mathbf U_{T-2}^\top\\
        &=\lambda\beta_{T-1}\mathbf \Sigma_{T-2}\big((1-\beta_{T-1})\mathbf \Sigma_{T-2}+\beta_{T-1}\mathbf I\big)^{-1} + (1-\beta_{T-1})\mathbf{\Sigma}_{T-2}\big((1-\beta_{T-1})\mathbf \Sigma_{T-2}+\beta_{T-1}\mathbf I\big)^{-1}  (\lambda\beta_T+\alpha_T)\mathbf \Sigma_{T-1}\\
        &~~~~~~~~~~~~~~~~\cdot\big((1-\beta_{T-1})\mathbf \Sigma_{T-2}+\beta_{T-1}\mathbf I\big)^{-\top}\mathbf{\Sigma}_{T-2}^\top \\
        &=\mathbf \Sigma_{T-2}\big(\lambda\beta_{T-1}\mathbf \Sigma_{T-1}^{-1} + (1-\beta_{T-1}) (\lambda + (1-\lambda)\alpha_T)\mathbf \Sigma_{T-1}^{-1}\mathbf \Sigma_{T-2}\big)  \\
        &=\mathbf \Sigma_{T-2}\big(\lambda\mathbf \Sigma_{T-1}^{-1}(\beta_{T-1}\mathbf I + (1-\beta_{T-1})\mathbf \Sigma_{T-2} -\alpha_{T-1}\alpha_{T}\mathbf \Sigma_{T-2}) + \alpha_{T-1}\alpha_T\mathbf \Sigma_{T-1}^{-1}\mathbf \Sigma_{T-2}\big)\\
        &=\mathbf \Sigma_{T-2}\big(\lambda\mathbf \Sigma_{T-1}^{-1}\mathbf \Sigma_{T-1} + (1-\lambda)\alpha_{T-1}\alpha_T\mathbf \Sigma_{T-1}^{-1}\mathbf \Sigma_{T-2}\big)\\
        &=\mathbf \Sigma_{T-2}\big(\lambda\mathbf I+(1-\lambda)\alpha_{T}\alpha_{T-1}\mathbf \Sigma_{T-1}^{-1}\mathbf \Sigma_{T-2}\big), ~~~\text{as } \bar{\alpha}_T\to 0.
    \end{align*}
\end{small}

    Now, let's prove the general case by induction. For $0\leq k\leq T-3$, suppose that the variance of $p_{\lambda}(\mathbf x_{k+1}|\mathbf y)$ is 
    \begin{align*}
        \mathbf \Sigma^{\lambda}_{k+1}=\mathbf \Sigma_{k+1}\Big(\lambda \mathbf I+(1-\lambda)\alpha_{k+2}\alpha_{k+3}\cdots\alpha_T\mathbf \Sigma_{T-1}^{-1}\mathbf \Sigma_{k+1}\Big),
    \end{align*}
    and the expectation is 
    \begin{align*}
        \boldsymbol\mu_{k+1}^{\lambda} &= \Big(\mathbf U_{k+1}\big(\mathbf U_{k+2}(\cdots (\mathbf U_{T-2}\mathbf V_{T-1}+\mathbf V_{T-2})\cdots)+\mathbf V_{k+2}\big)+\mathbf V_{k+1}\Big)\mathbb E_{p(\mathbf x_0|\mathbf y)}[X_0]\\
        &= (1-\alpha_{k+2}\alpha_{k+3}\cdots\alpha_T)\sqrt{\bar{\alpha}_{k+1}}\big((1-\bar{\alpha}_T)\mathbf I+\bar{\alpha}_T\text{Cov}_{p(\mathbf x_0|\mathbf y)}[X_0]\big)^{-1}\E_{p(\mathbf x_0|\mathbf y)}[X_0]\\
        &= (1-\alpha_{k+2}\alpha_{k+3}\cdots\alpha_T)\sqrt{\bar{\alpha}_{k+1}}\mathbb E_{p(\mathbf x_0|\mathbf y)}[X_0], ~~~\text{as } \bar{\alpha}_T\to 0.
    \end{align*}

    By Lemma \ref{Gaussian_lem} and $p_{\lambda}(\mathbf x_{k}|\mathbf x_{k+1},\mathbf y)=\mathcal{N}\Big(\mathbf U_{k}\mathbf x_{k+1}+\mathbf V_{k}\E_{p(\mathbf x_0|\mathbf y)}[X_0], ~\lambda \mathbf C_k\Big)$, we have mean and variance of $p_{\lambda}(\mathbf x_{k}|\mathbf y)$ as
\begin{small}
    \begin{align*}
        \boldsymbol{\mu}_{k}^{\lambda} &= \Big(\mathbf U_{k}(1-\alpha_{k+2}\alpha_{k+3}\cdots\alpha_T)\sqrt{\bar{\alpha}_{k+1}}\big((1-\bar{\alpha}_T)\mathbf I+\bar{\alpha}_T\text{Cov}_{p(\mathbf x_0|\mathbf y)}[X_0]\big)^{-1} + \mathbf V_{k}\Big)E_{p(\mathbf x_0|\mathbf y)}[X_0]\\
        &=\sqrt{\bar{\alpha}}_k
        \Big((1-\alpha_{k+2}\alpha_{k+3}\cdots\alpha_T)\big((1-\bar{\alpha}_{k})\mathbf I+\bar{\alpha}_{k}\text{Cov}_{p(\mathbf x_0|\mathbf y)}[X_0]\big)+(1-\alpha_{k+1})\big((1-\bar{\alpha}_T)\mathbf I+\bar{\alpha}_T\text{Cov}_{p(\mathbf x_0|\mathbf y)}[X_0]\big)\Big)\\
        &~~~~~~~~~~~~~~~~~~~~~\cdot\big((1-\bar{\alpha}_{k+1})\mathbf I+\bar{\alpha}_{k+1}\text{Cov}_{p(\mathbf x_0|\mathbf y)}[X_0]\big)^{-1}\big((1-\bar{\alpha}_T)\mathbf I+\bar{\alpha}_T\text{Cov}_{p(\mathbf x_0|\mathbf y)}[X_0]\big)^{-1} \E_{p(\mathbf x_0|\mathbf y)}[X_0] \\
        &=\sqrt{\bar{\alpha}}_k
        \Big((1-\bar{\alpha}_{k+1})(1-\alpha_{k+1}\alpha_{k+2}\cdots\alpha_{T})\mathbf{I}+(1-\alpha_{k+1}\alpha_{k+2}\cdots\alpha_{T})\bar{\alpha}_{k+1}\Cov_{p(\mathbf x_0|\mathbf y)}[\mathbf x_0]\Big)\\
        &~~~~~~~~~~~~~~~~~~~~~\cdot\big((1-\bar{\alpha}_{k+1})\mathbf I+\bar{\alpha}_{k+1}\text{Cov}_{p(\mathbf x_0|\mathbf y)}[X_0]\big)^{-1}\big((1-\bar{\alpha}_T)\mathbf I+\bar{\alpha}_T\text{Cov}_{p(\mathbf x_0|\mathbf y)}[X_0]\big)^{-1} \E_{p(\mathbf x_0|\mathbf y)}[X_0] \\
        &=(1-\alpha_{k+1}\alpha_{k+2}\cdots\alpha_T)\sqrt{\bar{\alpha}_k}\big((1-\bar{\alpha}_T)\mathbf I+\bar{\alpha}_T\text{Cov}_{p(\mathbf x_0|\mathbf y)}[X_0]\big)^{-1}\mathbb E_{p(\mathbf x_0|\mathbf y)}[X_0]\\
        &=(1-\alpha_{k+1}\alpha_{k+2}\cdots\alpha_T)\sqrt{\bar{\alpha}_k}\mathbb E_{p(\mathbf x_0|\mathbf y)}[X_0] ~~~\text{as } \bar{\alpha}_T\to 0, 
    \end{align*}
    \begin{align*}
        \mathbf \Sigma_{k}^{\lambda}&=\lambda \mathbf C_k+\mathbf U_{k}\mathbf \Sigma^{\lambda}_{k+1}\mathbf U_{k}^{\top}\\
        &=\lambda\beta_{k+1}\mathbf \Sigma_k\big((1-\beta_{k+1})\mathbf \Sigma_k+\beta_{k+1}\mathbf I\big)^{-1} +(1-\beta_{k+1})\mathbf \Sigma_k\big((1-\beta_{k+1})\mathbf \Sigma_k+\beta_{k+1}\mathbf I\big)^{-1}\\
        &~~~~~~~~~~~~~~\cdot \mathbf \Sigma_{k+1}\big(\lambda \mathbf I+(1-\lambda)\alpha_{k+2}\alpha_{k+3}\cdots\alpha_T\mathbf \Sigma_{T-1}^{-1}\mathbf \Sigma_{k+1}\big) \big((1-\beta_{k+1})\mathbf \Sigma_k+\beta_{k+1}\mathbf I\big)^{-\top}\mathbf \Sigma_k^{\top}\\
        &=\mathbf \Sigma_k \Big(\lambda\beta_{k+1}\mathbf{\Sigma}_{k+1}^{-1}+(1-\beta_{k+1})\big(\lambda \mathbf{\Sigma}_{k+1}^{-1}\mathbf \Sigma_k^{\top}+(1-\lambda)\alpha_{k+2}\alpha_{k+3}\cdots\alpha_T\mathbf \Sigma^{-1}_{T-1}\mathbf \Sigma_k^{\top}\big)\Big)\\
        &=\mathbf \Sigma_k \Big(\lambda\mathbf{\Sigma}_{k+1}^{-1}\big(\beta_{k+1}\mathbf{I}+(1\!\!-\!\!\beta_{k+1})\mathbf \Sigma_k\!-\!\alpha_{k+1}\alpha_{k+2}\alpha_{k+3}\cdots\alpha_{T}\mathbf \Sigma_{k+1}\mathbf \Sigma^{-1}_{T-1}\mathbf \Sigma_{k}\big)+\alpha_{k+1}\alpha_{k+2}\alpha_{k+3}\cdots\alpha_{T}\mathbf \Sigma^{-1}_{T-1}\mathbf \Sigma_{k}\Big)\\
        &=\mathbf \Sigma_k\Big(\lambda\big(\mathbf{I} - \alpha_{k+1}\cdots\alpha_T\mathbf \Sigma_{T-1}^{-1}\mathbf \Sigma_{k} \big)+\alpha_{k+1}\cdots\alpha_T\mathbf \Sigma_{T-1}^{-1}\mathbf \Sigma_{k}\Big) \\
        &=\mathbf \Sigma_{k}\Big(\lambda \mathbf I+(1-\lambda)\alpha_{k+1}\alpha_{k+2}\cdots\alpha_T\mathbf \Sigma_{T-1}^{-1}\mathbf \Sigma_{k}\Big).
    \end{align*}
\end{small}
    In particular, when $\bar\alpha_T\to 0$, the variance of $p_{\lambda}(\mathbf x_0|\mathbf y)$ is 
    \begin{align*}
        \mathbf \Sigma_{0}^{\lambda}&=\mathbf \Sigma_{0}\Big(\lambda \mathbf I+(1-\lambda)\bar{\alpha}_T\mathbf \Sigma_{T-1}^{-1}\mathbf \Sigma_{0}\Big)
        \to 
        \lambda\text{Cov}_{p(\mathbf x_0|\mathbf y)}[X_0],
    \end{align*}
    and the mean is
    \begin{align*}
        \boldsymbol{\mu}_0^{\lambda}=(1-\bar{\alpha}_T)\sqrt{\bar{\alpha}_0}\big((1-\bar{\alpha}_T)\mathbf I+\bar{\alpha}_T\text{Cov}_{p(\mathbf x_0|\mathbf y)}[X_0]\big)\mathbb E_{p(\mathbf x_0|\mathbf y)}[X_0]\to \mathbb E_{p(\mathbf x_0|\mathbf y)}[X_0].
    \end{align*}

    \section{Proof of Theorem \ref{thm_optimality_condi_Gaussian}} \label{Appendix_Proof_Optimality}
    
    \underline{Optimality:}
    First, we shall show that there is no loss of optimality in assuming that $\hat{X}$ is jointly Gaussian with $X$ given $\mathbf y$. Let $\hat{X}_G$ be a random variable with the
    same first and second-order statistics as $\hat{X}$, and $p_{\hat{X}_G|Y}(\hat{\mathbf x}_G|\mathbf y)$ be a Gaussian distribution, i.e., $p_{\hat{X}_G|Y}(\hat{\mathbf x}_G|\mathbf y)\sim\mathcal{N}(\boldsymbol{\hat\mu}_{\mathbf y}, \hat{\mathbf \Sigma}_{\mathbf y})$. Since the first and second-order statistics are the same, we have $\E[||X-\hat{X}||^2]=\E[||X-\hat{X}_G||^2]$. Meanwhile, by \cite[Proposition 1.6.5]{Invitation_Wasserstein_space}, $W_2^2(p_{X|Y}(\mathbf x|\mathbf y),p_{\hat{X}|Y}(\hat{\mathbf x}|\mathbf y))\geq ||\boldsymbol\mu_{\mathbf y}-\boldsymbol{\hat\mu}_{\mathbf y}||^2_2+\Tr(\mathbf \Sigma_{\mathbf y}+\hat{\mathbf \Sigma}_{\mathbf y}-2\big(\mathbf \Sigma_{\mathbf y}^{\frac{1}{2}}\hat{\mathbf \Sigma}_{\mathbf y}{\mathbf \Sigma}_{\mathbf y}^{\frac{1}{2}}\big)^{\frac{1}{2}})=W_2^2(p_{X|Y}(\mathbf x|\mathbf y),p_{\hat{X}_G|Y}(\hat{\mathbf x}_G|\mathbf y))$, where $W_2(p,q)$ denotes the Wasserstein-2 (W2) distance between two distributions $p$ and $q$.
    
    Thus, we can assume that the construction $\hat{X}$ is jointly Gaussian with $X$ given $\mathbf y$. Together with the Markov chain $X-Y-\hat{X}$, i.e., $p_{X,\hat{X}|Y}(\mathbf x,\hat{\mathbf x}|\mathbf y)=p_{X|Y}(\mathbf x|\mathbf y)p_{\hat X|Y}(\hat{\mathbf x}|\mathbf y)$, the optimization problem \eqref{DP_optimal} in Theorem \ref{thm_optimality_condi_Gaussian} becomes
    \begin{align*}
        D(P) = \min_{\boldsymbol{\hat\mu}_{\mathbf y}, \hat{\mathbf \Sigma}_{\mathbf y}}&||\boldsymbol\mu_{\mathbf y}-\boldsymbol{\hat\mu}_{\mathbf y}||^2_2+\Tr(\mathbf \Sigma_{\mathbf y})+\Tr(\hat{\mathbf \Sigma}_{\mathbf y})\\
        \text{s.t.}~ &||\boldsymbol\mu_{\mathbf y}-\boldsymbol{\hat\mu}_{\mathbf y}||^2_2+\Tr\big(\mathbf \Sigma_{\mathbf y}+\hat{\mathbf \Sigma}_{\mathbf y}-2\big(\mathbf \Sigma_y^{\frac{1}{2}}\hat{\mathbf \Sigma}_{\mathbf y}\mathbf \Sigma_{\mathbf y}^{\frac{1}{2}}\big)^{\frac{1}{2}}\big)\leq P^2.
    \end{align*}
    
    Without loss of optimality, we set $\boldsymbol{\hat\mu}_{\mathbf y}=\boldsymbol\mu_{\mathbf y}$. Consider the KKT condition with dual variable $\nu$:
    \begin{align}
        &\nabla_{\hat{\mathbf \Sigma}_{\mathbf y}} \Big(\Tr(\mathbf \Sigma_{\mathbf y})+\Tr(\hat{\mathbf \Sigma}_{\mathbf y}) + \nu \big( \Tr\big(\mathbf \Sigma_{\mathbf y}+\hat{\mathbf \Sigma}_{\mathbf y}-2\big(\mathbf \Sigma_{\mathbf y}^{\frac{1}{2}}\hat{\mathbf \Sigma}_{\mathbf y}\mathbf \Sigma_{\mathbf y}^{\frac{1}{2}}\big)^{\frac{1}{2}}\big) \big)\Big) = \mathbf I + \nu \mathbf I -\nu\mathbf \Sigma_{\mathbf y}^{\frac{1}{2}}\big(\mathbf \Sigma_{\mathbf y}^{\frac{1}{2}}\hat{\mathbf \Sigma}_{\mathbf y}\mathbf \Sigma_{\mathbf y}^{\frac{1}{2}}\big)^{-\frac{1}{2}}\mathbf \Sigma_{\mathbf y}^{\frac{1}{2}} = 0, \label{KKT1}\\
        &\nu \big(\Tr\big(\mathbf \Sigma_{\mathbf y}+\hat{\mathbf \Sigma}_{\mathbf y}-2\big(\mathbf \Sigma_y^{\frac{1}{2}}\hat{\mathbf \Sigma}_{\mathbf y}\mathbf \Sigma_{\mathbf y}^{\frac{1}{2}}\big)^{\frac{1}{2}}\big) - P^2 \big) = 0\label{KKT2},\\
        &\nu\geq 0.
    \end{align}
    
    With \eqref{KKT1}, we have $\hat{\mathbf \Sigma}_{\mathbf y}=\big(\frac{\nu}{1+\nu}\big)^2{\mathbf \Sigma}_{\mathbf y}$. Plugging in \eqref{KKT2}, we have 
    \begin{align*}
        \nu\Big(\Tr({\mathbf \Sigma}_{\mathbf y}) + \big(\frac{\nu}{1+\nu}\big)^2\Tr(\mathbf \Sigma_{\mathbf y}) - 2\big(\frac{\nu}{1+\nu}\big)\Tr\big(\big(\mathbf \Sigma_{\mathbf y}^{\frac{1}{2}}\mathbf \Sigma_{\mathbf y}\mathbf \Sigma_{\mathbf y}^{\frac{1}{2}}\big)^{\frac{1}{2}}\big)-P^2\Big) = \nu\Big(\frac{1}{(1+\nu)^2}\Tr(\mathbf \Sigma_{\mathbf y})-P^2\Big)=0.
    \end{align*}
    When $P> \sqrt{\Tr(\mathbf \Sigma_{\mathbf y})}$, $\nu$ should be zero. When $P\leq \Tr(\mathbf \Sigma_{\mathbf y})$, we have $\nu=\sqrt{\frac{\Tr(\mathbf \Sigma_{\mathbf y})}{P^2}}-1$, and the distortion level is 
    \begin{align*}
        D(P) = \Tr(\mathbf \Sigma_{\mathbf y}) + \big(\frac{\nu}{\nu+1}\big)^2\Tr(\mathbf \Sigma_{\mathbf y}) = \Big(1+\big(1-\sqrt{\frac{P^2}{\Tr(\mathbf \Sigma_{\mathbf y})}}\big)^2\Big)\Tr(\mathbf \Sigma_{\mathbf y})=\Tr(\mathbf \Sigma_{\mathbf y})+\big(\sqrt{\Tr(\mathbf \Sigma_{\mathbf y})}-P\big)^2.
    \end{align*}
    
    In summary, the optimal conditional distortion-perception tradeoff with MSE and W2 constraint is 
    \begin{align}
        D(P)=\begin{cases}
            \Tr(\mathbf \Sigma_{\mathbf y})+\big(\sqrt{\Tr(\mathbf \Sigma_{\mathbf y})}-P\big)^2, ~\text{for } P\leq \sqrt{\Tr(\mathbf \Sigma_{\mathbf y})}\\
            \Tr(\mathbf \Sigma_{\mathbf y}),~\text{for }  P>\sqrt{\Tr(\mathbf \Sigma_{\mathbf y})}. 
        \end{cases}\tag{\ref{DP_optimal}}
    \end{align}
    
    \underline{Achievability:} In Theorem \ref{thm_reverse_mean_var}, we have shown that when $\bar\alpha_T\to 0$, the output distribution $p_{\lambda}(\mathbf x_0|\mathbf y)$ of the proposed reverse diffusion process \eqref{Eq_joint_inference_dist} is multivariate Gaussian with variance  
    \begin{align*}
        \mathbf \Sigma_{0}^{\lambda}&=\mathbf \Sigma_{0}\Big(\lambda \mathbf I+(1-\lambda)\bar{\alpha}_T\mathbf \Sigma_{T-1}^{-1}\mathbf \Sigma_{0}\Big)\to  \lambda\text{Cov}_{p(\mathbf x_0|\mathbf y)}[X_0] = \lambda \mathbf \Sigma_{\mathbf y},
    \end{align*}
    and mean
    \begin{align*}
        \boldsymbol\mu_0^{\lambda}=(1-\bar{\alpha}_T)\sqrt{\bar{\alpha}_0}\big((1-\bar{\alpha}_T)\mathbf I+\bar{\alpha}_T\text{Cov}_{p(\mathbf x_0|\mathbf y)}[X_0]\big)\mathbb E_{p(\mathbf x_0|\mathbf y)}[X_0]\to \mathbb E_{p(\mathbf x_0|\mathbf y)}[X_0] = \boldsymbol{\mu}_{\mathbf y}.
    \end{align*}
    Denote the reconstruction associated with $\lambda$ as $X_0^{\lambda}$ for $0\leq \lambda\leq 1$, and $p_{X_0^{\lambda}|Y}(\mathbf x_0^{\lambda}|\mathbf y)\triangleq p_{\lambda}(\mathbf x_0^{\lambda}|\mathbf y)$. Since both $p_{X_0^{\lambda}|Y}(\mathbf x_0^{\lambda}|\mathbf y)$ and $p_{X|Y}(\mathbf x|\mathbf y)$ are Gaussian, the Wasserstein-2 distance for two conditional distributions can be computed as
    \begin{align*}
        W_2^2(p_{X|Y}(\mathbf x|\mathbf y), p_{X_0^{\lambda}|Y}(\mathbf x_0^{\lambda}|\mathbf y)) &=\Tr(\mathbf \Sigma_{\mathbf y})+\lambda \Tr(\mathbf \Sigma_{\mathbf y})-2\sqrt{\lambda}\Tr((\mathbf \Sigma_{\mathbf y}^{\frac{1}{2}}\mathbf \Sigma_{\mathbf y}\mathbf \Sigma_{\mathbf y}^{\frac{1}{2}})^{\frac{1}{2}})\\
        &=(1-\sqrt{\lambda})^2\Tr(\mathbf \Sigma_{\mathbf y}).
    \end{align*}
    
    For the distortion, we have
    \begin{align*}
        \mathbb E_{p_{X_{0}^{\lambda},X|Y}(\mathbf x_0^{\lambda}, \mathbf x|\mathbf y)}[||X_0^{\lambda}-X||]&=\mathbb E_{p_{X_{0}^{\lambda},X|Y}(\mathbf x_0^{\lambda}, \mathbf x|\mathbf y)}[||X||^2 + ||X_0^{\lambda}||^2 - 2XX_0^{\lambda}]\\
        &=\mathbb E_{p_{X|Y}(\mathbf x|\mathbf y)}[||X||^2] + \mathbb E_{p_{X_0^{\lambda}|Y}(\mathbf x_0^{\lambda}|y)}[||X_0^{\lambda}||^2] - 2\mathbb E_{p_{X_0^{\lambda}|Y}(\mathbf x_0^{\lambda}|y)p_{X|Y}(\mathbf x|\mathbf y)}[XX_0^{\lambda}]\\
        &=\boldsymbol\mu_{\mathbf y}^{\top}\boldsymbol \mu_{\mathbf y} + \Tr(\mathbf \Sigma_y) + \boldsymbol\mu_{\mathbf y}^{\lambda\top}\boldsymbol\mu_{\mathbf y}^{\lambda} +\Tr(\mathbf \Sigma_{y}^{\lambda}) -2\boldsymbol\mu_{\mathbf y}^{\top}\boldsymbol\mu_{\mathbf y}^{\lambda}\\
        &=(1+\lambda)\Tr(\mathbf \Sigma_{y}^{\lambda}).
    \end{align*}
    Thus, the conditional distortion-perception tradeoff given by the scaled reverse diffusion process \eqref{Eq_joint_inference_dist} is
    \begin{align*}
        &D(\lambda) = (1+\lambda)\Tr(\mathbf \Sigma_y),\\
        &P^2(\lambda) = (1-\sqrt{\lambda})^2\Tr(\mathbf \Sigma_y),
    \end{align*}
    which by eliminating $\lambda$ is equivalent to
    \begin{align*}
        D(P)=\Tr(\mathbf \Sigma_{\mathbf y})+\big(\sqrt{\Tr(\mathbf \Sigma_{\mathbf y})}-P\big)^2, ~\text{for } P\leq \sqrt{\Tr(\mathbf \Sigma_{\mathbf y})}.
    \end{align*}
    Hence, the achieved tradeoff coincides with the optimal tradeoff \eqref{DP_optimal}.

    \section{Derivation of Mixture Gaussian Example}\label{Appendix_MixtureGaussian}
    Consider the mixture Gaussian distribution $X_0\sim p(x_0)$ with two components, where
    \begin{align*}
        p(x_0)=w_1\underbrace{\mathcal{N}(\mu, \sigma_1^2)}_{p_1(x_0)}+w_2\underbrace{\mathcal{N}(\mu, \sigma_2^2)}_{p_2(x_0)}.
    \end{align*} 
    The noisy observation is obtained by $Y=aX_0+\sigma_0\epsilon$, where $\epsilon\sim\mathcal{N}(0,1)$, i.e., $p(y|x_0)=\mathcal{N}(ax_0, \sigma_0^2)$. The joint distribution of $(Y, X_0)$ is 
    \begin{align*}
        p(y,x_0)=p(y|x_0)p(x_0)=w_1\underbrace{\mathcal{N}\bigg(\begin{bmatrix}
            y,\\
            x_0
        \end{bmatrix};
        \begin{bmatrix}
            a\mu_1,\\
            \mu_1
        \end{bmatrix},
        \begin{bmatrix}
            a^2\sigma_1^2+\sigma_0^2, &a\sigma_1^2\\
            a\sigma_1^2, &\sigma_1^2
        \end{bmatrix}\bigg)}_{f_1(x_0,y)} + w_2\underbrace{\mathcal{N}\bigg(\begin{bmatrix}
            y,\\
            x_0
        \end{bmatrix};
        \begin{bmatrix}
            a\mu_2,\\
            \mu_2
        \end{bmatrix},
        \begin{bmatrix}
            a\sigma_2^2+\sigma_0^2, &a\sigma_2^2\\
            a\sigma_2^2, &\sigma_2^2
        \end{bmatrix}\bigg)}_{f_2(x_0,y)}.
    \end{align*}
    Then the marginal distribution of $Y$ is 
    \begin{align*}
        p(y)=w_1\underbrace{\mathcal{N}(a\mu_1,a^2\sigma_1^2+\sigma_0^2)}_{p_1(y)}+w_2\underbrace{\mathcal{N}(a\mu_2,a^2\sigma_2^2+\sigma_0^2)}_{p_2(y)}.
    \end{align*}
    
    For component $f_1(x_0,y)$, it is a bivariate Gaussian distribution with marginals as $p_1(x_0)=\mathcal{N}(\mu_1,\sigma_1^2)$, and $p_1(y)=\mathcal{N}(a\mu_1,a^2\sigma_1^2+\sigma_0^2)$, with correlation $\rho=\frac{a\sigma_1}{\sqrt{a^2\sigma_1^2+\sigma_0^2}}$.
    
    Then $f_1(x_0,y)$ can be written as 
    \begin{align*}
        f_1(x_0,y) &= \frac{1}{2\pi\sigma_0\sigma_1}\exp\Big( -\frac{a^2\sigma_1^2+\sigma_0^2}{2\sigma_0^2}\Big[\big(\frac{x-\mu_1}{\sigma_1}\big)^2-2\frac{a\sigma_1}{\sqrt{a^2\sigma_1^2+\sigma_0^2}}\big(\frac{x-\mu_1}{\sigma_1}\big)\big(\frac{y-a\mu_1}{\sqrt{a^2\sigma_1^2+\sigma_0^2}}\big)+\big(\frac{y-a\mu_1}{\sqrt{a^2\sigma_1^2+\sigma_0^2}}\big)^2\Big] \Big)\\
        &=p_1(x_0|y)p_1(y),
    \end{align*}
    where $p_1(x_0|y)=\mathcal{N}\big((\frac{\mu_1}{\sigma_1^2}+\frac{ay}{\sigma_0^2})/(\frac{a^2}{\sigma_0^2}+\frac{1}{\sigma_1^2}), 1/(\frac{a^2}{\sigma_0^2}+\frac{1}{\sigma_1^2})\big)$ and $p_1(y)=\mathcal{N}(a\mu_1,a\sigma_1^2+\sigma_0^2)$. Similarly, we can write $f_2(x_0,y)$ as $p_2(x_0|y)p_2(y)$ where, $p_2(x_0|y)=\mathcal{N}\big((\frac{\mu_2}{\sigma_2^2}+\frac{ay}{\sigma_0^2})/(\frac{a^2}{\sigma_0^2}+\frac{1}{\sigma_2^2}), 1/(\frac{a^2}{\sigma_0^2}+\frac{1}{\sigma_2^2})\big)$, and $p_1(y)=\mathcal{N}(a\mu_2,a\sigma_2^2+\sigma_0^2)$.
    
    Then, the posterior distribution of $x_0$ given $y$ can be computed as
    \begin{align*}
        p(x_0|y)&=\frac{p(x_0,y)}{p(y)}=\frac{w_1f_1(x_0,y)+w_2f_2(x_0,y)}{w_1p_1(y)+w_2p_2(y)}\\
        &=\frac{w_1p_1(x_0|y)p_1(y)+w_2p_2(x_0|y)p_2(y)}{w_1p_1(y)+w_2p_2(y)}\\
        &=\underbrace{\frac{w_1p_1(y)}{w_1p_1(y)+w_2p_2(y)}}_{a_1(y)}p_1(x_0|y) + \underbrace{\frac{w_2p_2(y)}{w_1p_1(y)+w_2p_2(y)}}_{a_2(y)}p_2(x_0|y)\\
        &=a_1(y)\mathcal{N}\Big(\frac{\frac{\mu_1}{\sigma_1^2}+\frac{ay}{\sigma_0^2}}{\frac{a^2}{\sigma_0^2}+\frac{1}{\sigma_1^2}}, \frac{1}{\frac{a^2}{\sigma_0^2}+\frac{1}{\sigma_1^2}}\Big) + a_2(y)\mathcal{N}\Big(\frac{\frac{\mu_2}{\sigma_2^2}+\frac{ay}{\sigma_0^2}}{\frac{a^2}{\sigma_0^2}+\frac{1}{\sigma_1^2}}, \frac{1}{\frac{a^2}{\sigma_0^2}+\frac{1}{\sigma_1^2}}\Big).
    \end{align*}
    
    Thus, the MMSE estimator is $\mathbb{E}_{p(x_0|y)}[X_0]=a_1(y)(\frac{\mu_1}{\sigma_1^2}+\frac{ay}{\sigma_0^2})/(\frac{a^2}{\sigma_0^2}+\frac{1}{\sigma_1^2})+a_2(y)(\frac{\mu_2}{\sigma_2^2}+\frac{ay}{\sigma_0^2})/(\frac{a^2}{\sigma_0^2}+\frac{1}{\sigma_1^2})$.

    \section{Experimental Details and More Experimental Results} \label{Appendix_more_exp}

    \subsection{Experimental Details}\label{Appendix_Exp_detail}
    \subsubsection{Two-dimensional datasets} Here, we list the architecture design and choices of hyperparameters for the two-dimensional datasets.

    \textbf{Network architecture:} We use a simple architecture modified from \cite{bortoli2021diffusion_Schrodinger_Bridge_Score}. For the score network, the input point $\mathbf x$ and the time index $k$ are fed to an MLP Block, respectively, where each MLP Block is a multilayer perceptron network. Then, we concatenate the outputs of two MLP Blocks and then feed the concatenated output into a third MLP Blocks. For PSCGAN, the generator of CGAN is also built upon MLP Blocks. Specifically, the noisy observation $\mathbf y$ and initial noise $\mathbf z$ are fed to an MLP Block respectively, and the concatenated output is fed to another MLP Block. The discriminator of CGAN involves five linear layers, and leaky Relu is used for the activation function. Note that the number of parameters for the generator and discriminator are 25682 and 25025, respectively. The total number of parameters for the score network is 26498.
    
    \textbf{Choices of hyperparameters:} We set $T = 1000$ and a linear schedule from $\beta_1 = 10^{-4}$ to $\beta_T = 0.02$. Meanwhile, $\tilde{\sigma}_{k}$ is set to be $\beta_k$. For pinwheel dataset, the $\zeta_{k,\lambda}$ is set to be $1.2+1.8\lambda$, and for S-curve and moon datasets, $\zeta_{k,\lambda}$ is set to be $1+\lambda$ for all $k=0,1,\cdots, T$. For PSCGAN, we follow the setup shown in the original paper \cite{PSCGAN}. 
    
    All experiments for two-dimensional datasets were conducted on a single NVIDIA RTX A6000 GPU.
    
    \subsubsection{FFHQ dataset}
    Here we list the choices of hyperparameters for the FFHQ dataset. Note that the score network for our sampling method was taken from \cite{chung2023DPS}, which was trained from scratch using 49k training data for 1M steps. The pre-trained model for PSCGAN is taken from the original paper \cite{PSCGAN}.
    
    \textbf{Choices of hyperparameters:} We set $T = 1000$ and a linear schedule from $\beta_1 = 10^{-4}$ to $\beta_T = 0.02$. Meanwhile, $\tilde{\sigma}_{k}$ is set to be $\beta_k\frac{1-\bar{\alpha}_{k-1}}{1-\bar{\alpha}_{k}}$. The choices of $\{\zeta_{k,\lambda}\}_{k=0}^T$ are heuristic and may be slightly different for different devices to get the best results. Recall that $\{\zeta_{k,\lambda}\}_{k=0}^T$ control the weight of the conditional score. Theoretically, if we directly follow Bayes' rule and set the weight of $\nabla_{\mathbf{x}_k}p(\mathbf{x}_k)$ and $\nabla_{\mathbf{x}_k}p(\mathbf y|\mathbf{x}_k)$ to be equal, we can obtain the theoretical value as $\zeta_{k}^{'}=\frac{1-\alpha_k}{2\sqrt{\alpha_k}\sigma_n^2}$. However, the choice of $\zeta_{k}^{'}$ is not practical. Since $s_{\theta}(\mathbf x_k,k)$ is usually much larger than $\hat{c}(\hat{\mathbf x}_0)$ in Algorithm 1, $\zeta_{k}^{'}$ is too small to reflect information on the conditional score properly. Thus, we still use the heuristic choices of $\zeta_{k,\lambda}$.

    In general, for small $\lambda$'s (e.g., $\leq 0.6$), the $\{\zeta_{k,\lambda}\}_{k=0}^T$ need to be set large to get good reconstruction, while for $\lambda$ close to 1, small $\{\zeta_{k,\lambda}\}_{k=0}^T$ leads to better images. Large $\{\zeta_{k,\lambda}\}_{k=0}^T$ for $\lambda>0.7$ would result in degraded reconstructions. The possible reason is that for small $\lambda$ (with less stochasticity), the conditional information becomes more important in constructing a good image, leading to a greater reliance on the conditional score. When $\lambda$ is large, too much conditional information may conflict with the great stochasticity. In this paper, we mainly focus on tuning $\{\zeta_{k,\lambda}\}_{k=0}^T$ as a function of $\lambda$. Thus, $\{\zeta_{k,\lambda}\}_{k=0}^T$ is a constant for all $k$ and $\mathbf y$. It is possible to further tune the parameters as a function of $k$ or $\|\mathbf{y}-\mathcal{A}(\mathbf{x}_0)\|_2^2$ \cite{chung2023DPS}. In practice, the choices in Table \ref{GS0.3_zeta} , \ref{GS0.5_zeta} and \ref{SR_zeta} could be considered for discrete $\lambda\in\{0,0.1,\cdots,1\}$.
    
    For PSCGAN and DiffPIR, we use the hyperparameters according to the suggested values in the respective papers. All experiments are conducted on a single NVIDIA A100 GPU.
    
    \begin{table}[!t]
        \center
        \caption{Choices of $\{\zeta_{k,\lambda}\}_{k=0}^T$ on Gaussian deblurring task with $\sigma_n=0.3$ for discrete $\lambda$'s.}\label{GS0.3_zeta}
        \begin{tabular}{ccccccccccc}
          \toprule
          $\lambda=0$ & $\lambda=0.1$ &  $\lambda=0.2$  & $\lambda=0.3$ & $\lambda=0.4$ & $\lambda=0.5$ & $\lambda=0.6$ &  $\lambda=0.7$  & $\lambda=0.8$ & $\lambda=0.9$ & $\lambda=1$  \\ 
          \midrule 
          39 & 24 & 24 & 26 & 26 & 40 & 22 & 18 & 12  & 12 & 6 \\
          \bottomrule
          \end{tabular}
      \end{table}

      \begin{table}[!t]
        \center
        \caption{Choices of $\{\zeta_{k,\lambda}\}_{k=0}^T$ on Gaussian deblurring task with $\sigma_n=0.5$ for discrete $\lambda$'s.}\label{GS0.5_zeta}
        \begin{tabular}{ccccccccccc}
          \toprule
          $\lambda=0$ & $\lambda=0.1$ &  $\lambda=0.2$  & $\lambda=0.3$ & $\lambda=0.4$ & $\lambda=0.5$ & $\lambda=0.6$ &  $\lambda=0.7$  & $\lambda=0.8$ & $\lambda=0.9$ & $\lambda=1$  \\ 
          \midrule 
          33 & 33 & 33 & 37 & 40 & 40 & 33 & 23 & 15  & 10 & 6.5 \\
          \bottomrule
          \end{tabular}
      \end{table}

      \begin{table}[!t]
        \center
        \caption{Choices of $\{\zeta_{k,\lambda}\}_{k=0}^T$ on super-resolution task with scale factor 8 for discrete $\lambda$'s.}\label{SR_zeta}
        \begin{tabular}{ccccccccccc}
          \toprule
          $\lambda=0$ & $\lambda=0.1$ &  $\lambda=0.2$  & $\lambda=0.3$ & $\lambda=0.4$ & $\lambda=0.5$ & $\lambda=0.6$ &  $\lambda=0.7$  & $\lambda=0.8$ & $\lambda=0.9$ & $\lambda=1$  \\ 
          \midrule 
          26 & 24 & 24 & 24 & 30 & 24 & 20 & 15 & 12 & 12 & 10 \\
          \bottomrule
          \end{tabular}
      \end{table}
    

    \subsection{More Experimental Results}\label{Appendix_Exp_More_results}
    \subsubsection{Two-dimensional datasets}
    We provide additional experiments on two-dimensional datasets, including more data distributions and validation of adjusting the variance scale.
    
    \textbf{More data distributions:} Other than pinwheel data points shown in Section \ref{sub-sec:2DExpriments}, we illustrate the results on S-curve and moon-type data distributions. Fig. \ref{Fig_recon_scurve_moon} shows the original distributions, noisy distributions, as well as the reconstructions for each dataset. The numerical DP tradeoffs are depicted in Fig. \ref{Fig_DP_scurve_moon}. Similar to the pinwheel case, our score-based method achieves a much larger range of tradeoffs compared to the GAN-based approach, revealing great effectiveness and optimality.
    \begin{figure}[!htpb]
        \centering
        \includegraphics[width=1\textwidth]{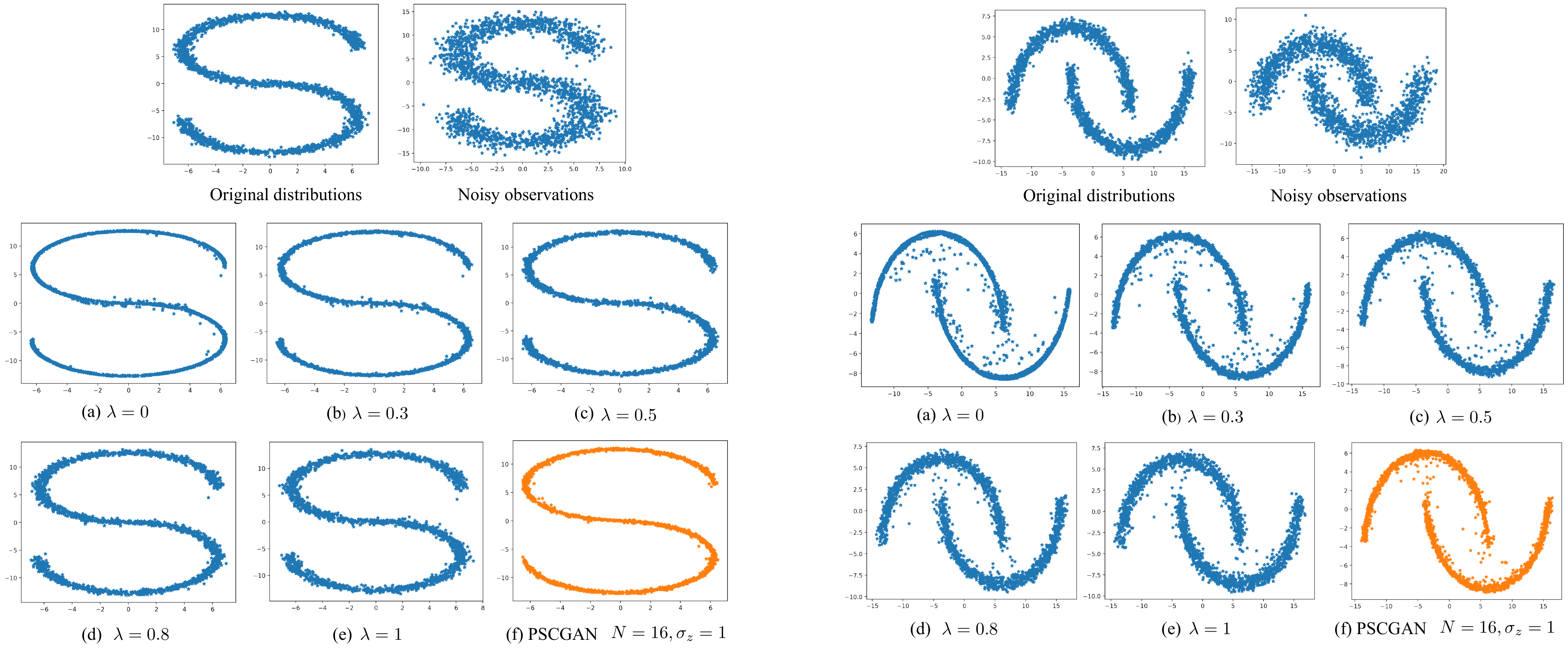}
        \caption{Experiments on the S-curve dataset (left) and Moon dataset (right). The first row illustrates the original distribution and the noisy observation $Y$, given by $Y=aX+N$ for $N\sim\mathcal{N}(0,\sigma_n^2\mathbf{I})$ (right). The second and third row shows the reconstructions on each dataset:  (a)-(e) variance-scaled reverse diffusion process with different $\lambda$'s; (f) PSCGAN with $N=16,\sigma_z=1$.}\label{Fig_recon_scurve_moon}
    \end{figure}
    
    \begin{figure}[!t]
        \centering
        \includegraphics[width=0.85\textwidth]{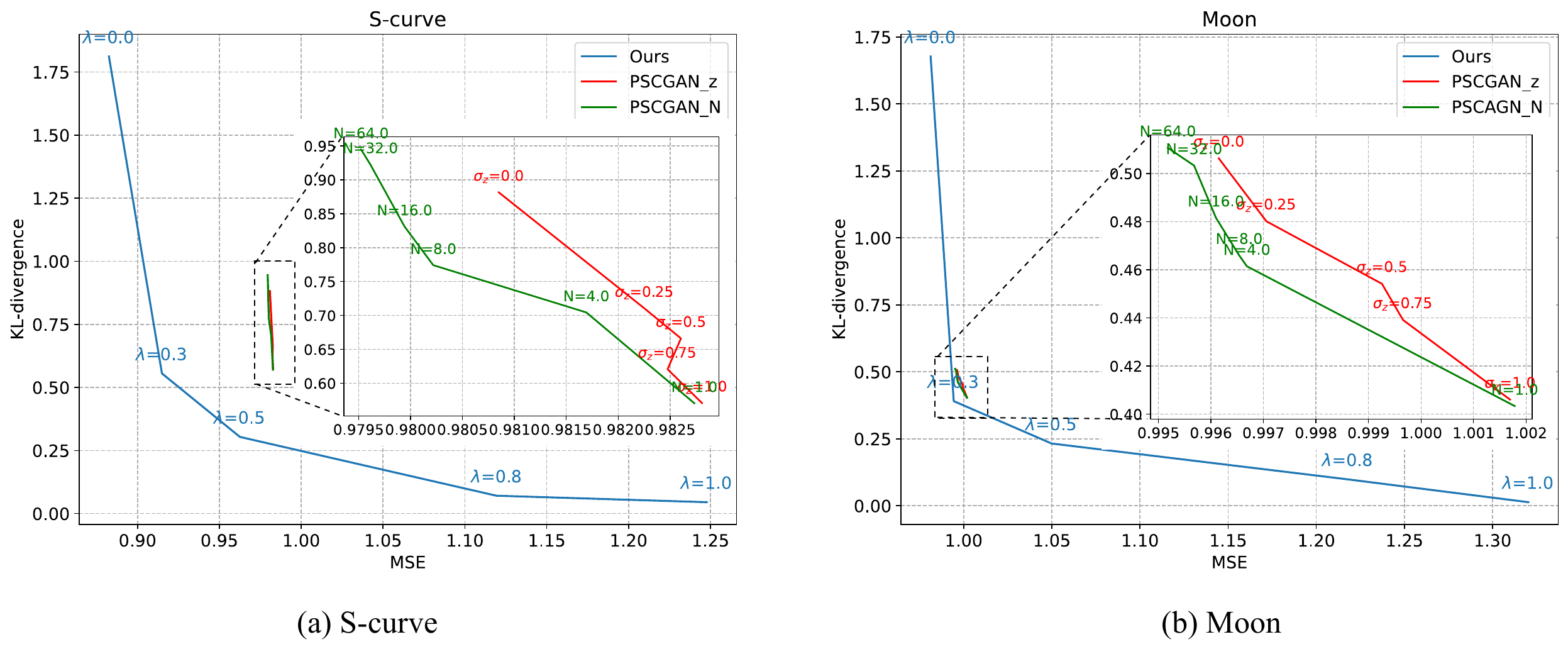}
        \caption{DP tradeoff on S-curve (left) and moon-type (right) datasets traversed by our variance-scaled reverse diffusion process and PSCGAN. }\label{Fig_DP_scurve_moon}
    \end{figure}

    \textbf{Validation of adjusting the variance scale:}
    In the original DPS sampling procedure \cite[Algorithm 1]{chung2023DPS}, there is a hyperparameter $\zeta_k$ controlling the weight that is given to the likelihood $\nabla_{\mathbf x_k}||y-\mathcal{A}(\hat{\mathbf x}_0(\mathbf x_k))||^2_2$, which may also affect the distortion-perception performance. Theorem \ref{thm_reverse_mean_var} and \ref{thm_reverse_mean_var} show that the proposed variance-scaled diffusion process serves as the optimal solution to the DP tradeoff for conditional multivariate Gaussian. In contrast, there is no theoretical guarantee that adjusting the DPS weight $\zeta_k$ in Algorithm 1 can traverse the optimal DP tradeoff. We conduct a simple experiment on the pinwheel dataset, which compares the performance of the proposed variance-scaled reverse diffusion process and the DPS sampling procedure with adjusted $\zeta_k$. Fig. \ref{Fig_DP_pinwheel_zeta} demonstrates that adjusting $\zeta_k$ for fixed $\lambda = 1$ is inferior to our variance-scaled method and unable to traverse the tradeoff.

    \begin{figure}[!t]
        \centering
        \includegraphics[width=0.4\textwidth]{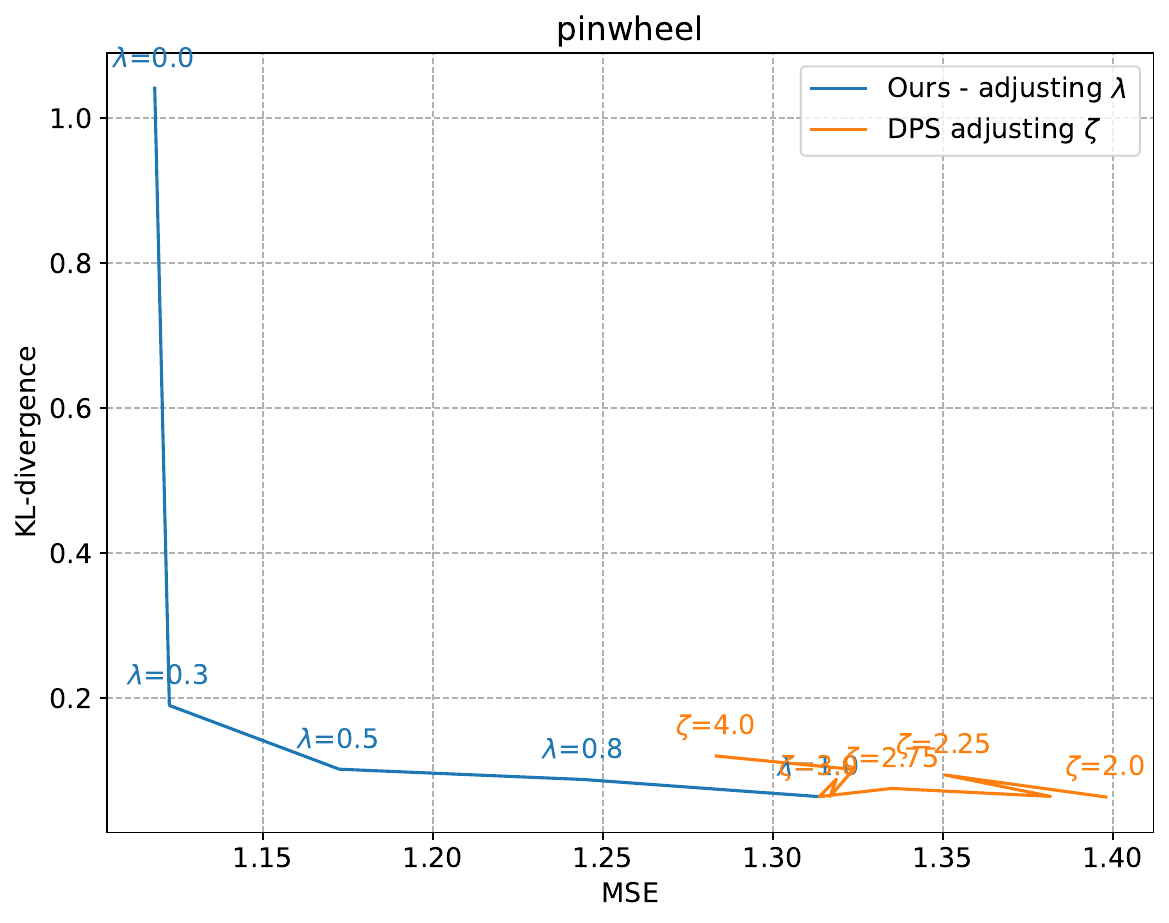}
        \caption{DP tradeoff on S-curve (left) and moon-type (right) datasets traversed by our variance-scaled reverse diffusion process and PSCGAN. }\label{Fig_DP_pinwheel_zeta}
    \end{figure}

    \subsubsection{FFHQ dataset}
    We provide more experimental results on the FFHQ dataset, including the effect of increasing stochasticity, more metrics, and more examples.

    \textbf{Increasing stochasticity:}
    It is observed in the mixture Gaussian example (Section \ref{sub-sec:mixture_gaussian}) that for $\lambda=0$, the trajectories are deterministic and converge to the MMSE point given an initial $x_T$. When $\lambda$ increases, the generated trajectories follow the form of the posterior distribution and show more stochasticity. This phenomenon can also be observed in real-world datasets. As shown in Fig. \ref{Fig_on1_img}, the reconstructions show more stochasticity with $\lambda$ increasing. Specifically, details such as hairs, eye expressions, and the shape of the mouth exhibit more variations. The images become sharper with the increase in MSE.
    
    \begin{figure}[!t]
        \centering
        \includegraphics[width=1\textwidth]{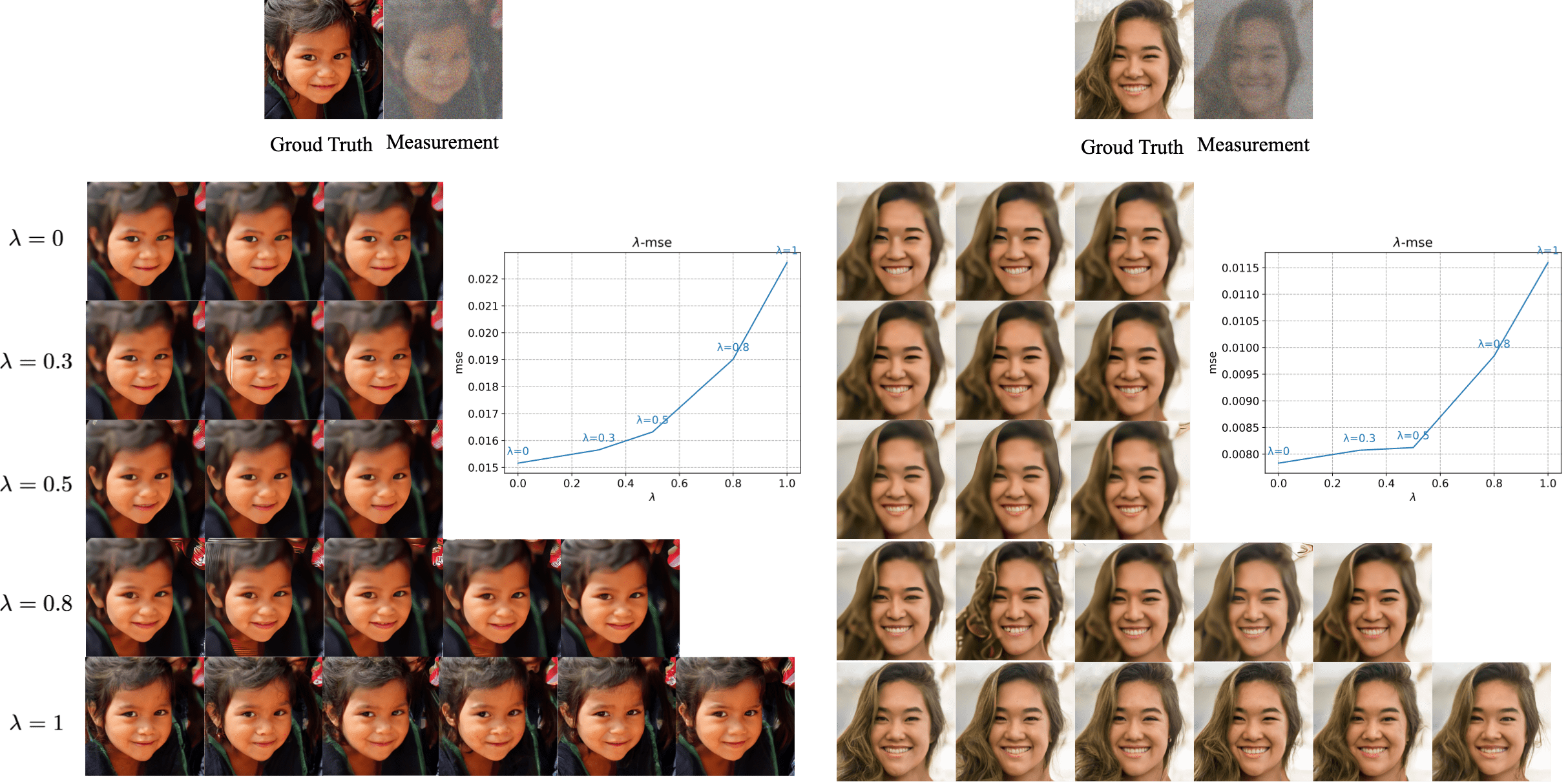}
        \caption{Multiple samples with different $\lambda$'s. As $\lambda$ increases, the reconstructions show more stochasticity, and the MSE increases.}\label{Fig_on1_img}
    \end{figure}

    \textbf{More metrics:} We report more metrics of Gaussian deblur task with additive noise of $\sigma_n = 0.3$ on FFHQ dataset, including PSNR for distortion and LPIPS \cite{LPIPS} for perception measure. It can be shown in Table \ref{LPIPS} that when $\lambda$ increases, PSNR becomes worse while LPIPS becomes better. This phenomenon coincides with the results of MSE and FID, indicating that the proposed method can effectively traverse the tradeoff between distortion and perception.

    \begin{table}[]
        \centering
        \begin{tabular}{ccccccccc}
          \toprule
        \multirow{2}{*}{Metrics} & \multicolumn{5}{c}{Ours} & \multicolumn{2}{c}{PSCGAN} & DiffPIR \\ \cline{2-9} 
         & \multicolumn{1}{l}{$\lambda=0$} & \multicolumn{1}{l}{$\lambda=0.3$} & \multicolumn{1}{l}{$\lambda=0.5$} & \multicolumn{1}{l}{$\lambda=0.8$} & \multicolumn{1}{l}{$\lambda=1$} & \multicolumn{1}{l}{$N=1$} & \multicolumn{1}{l}{$N=64$} &  \\
         \midrule
        PSNR$\uparrow$ & 25.27 & 24.93 & 24.80 & 24.47 & 24.40 & 22.10 & 24.39 & 22.73  \\
        LPIPS$\downarrow$ & 0.368 & 0.337 & 0.329 &	0.312 & 0.263 & 0.304 & 0.350 & 0.262 \\
        \bottomrule
        \end{tabular}
        \caption{Quantitative evaluation (PSNR, LPIPS) of Gaussian deblur task with additive noise of $\sigma_n=0.3$ on FFHQ dataset.}\label{LPIPS}
      \end{table}
    
    \textbf{More examples for different tasks:} Fig. \ref{Fig_FFHQ_GS_supp} shows more samples from the FFHQ dataset on the Gaussian deblurring task. We test the methods on different noise levels. Note that the PSCGAN is trained on $\sigma_n=0.3$. We can see that with a single score network, our method can robustly traverse DP on different noise levels. The PSCGAN trained on $\sigma_n=0.3$ fails to generate valid images when $\sigma_n=0.5$. More examples of the super-resolution task are shown in Fig. \ref{Fig_FFHQ_SR_supp}.

    \begin{figure}[t]
        \centering
        \includegraphics[width=0.7\textwidth]{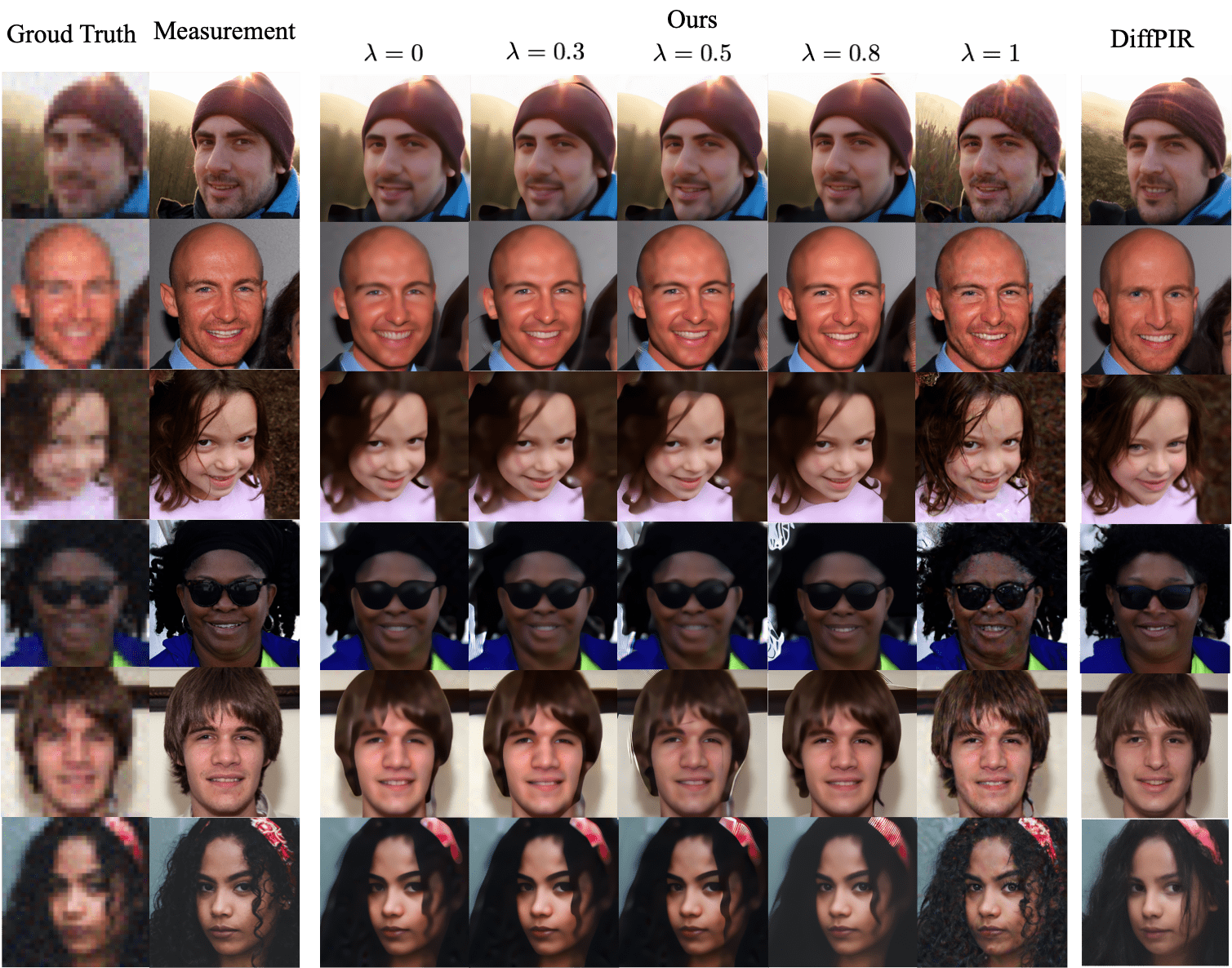}
        \caption{More examples on FFHQ dataset of super-resolution with downsampling scale 8.}\label{Fig_FFHQ_SR_supp}
    \end{figure}
    
    \begin{figure}[t]
        \centering
        \includegraphics[width=0.85\textwidth]{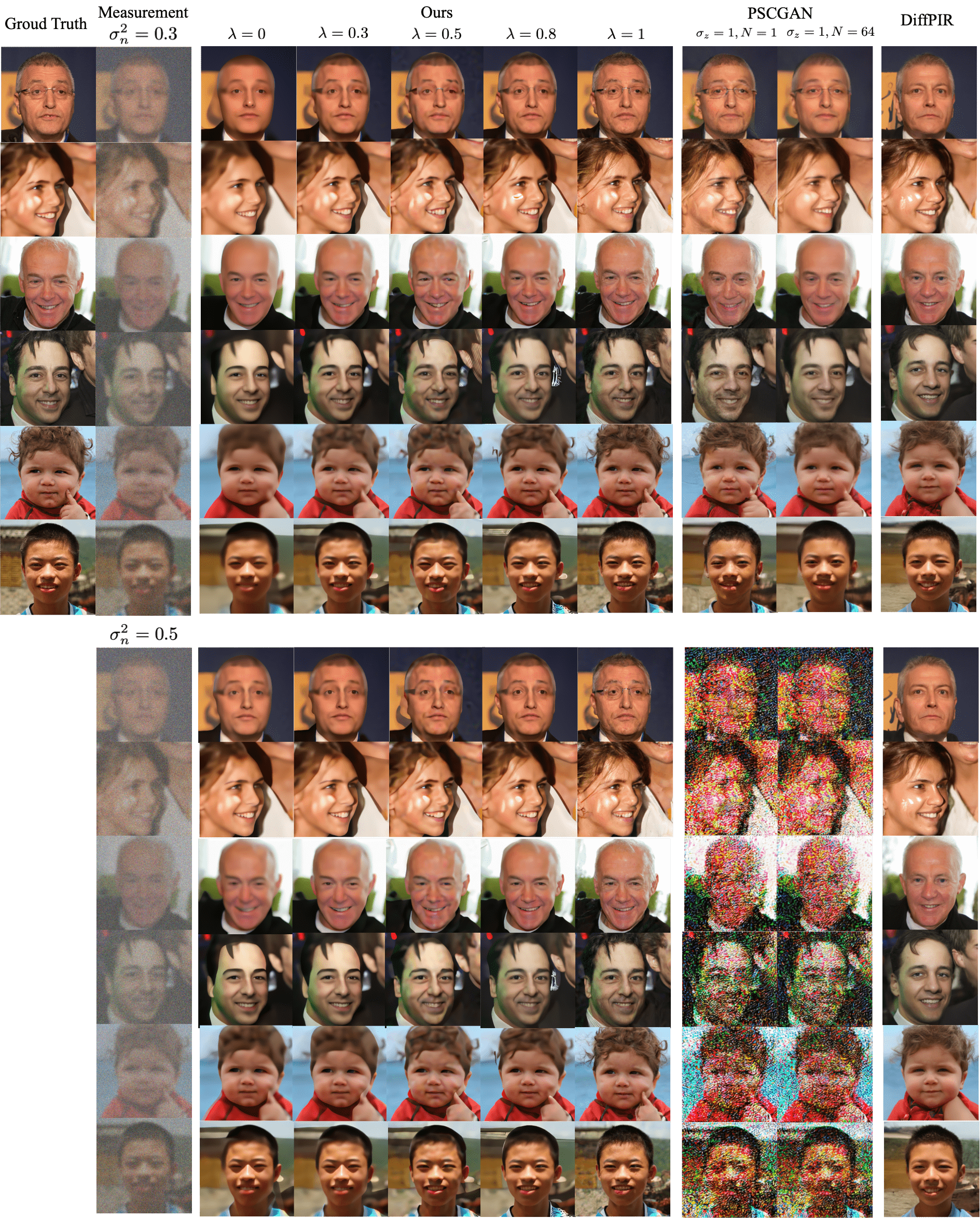}
        \caption{More examples on FFHQ dataset of Gaussian deblurring for both $\sigma_n=0.3$ and $\sigma_n=0.5$. Note that for each method, we use the same pre-trained model for both noise levels. With a single score network, our method can robustly traverse DP on different noise levels. The PSCGAN trained on $\sigma_n=0.3$ fails to generate valid images when $\sigma_n=0.5$.}\label{Fig_FFHQ_GS_supp}
    \end{figure}

}


\newpage
\bibliographystyle{IEEEtran}
\bibliography{refer}

\end{document}